\documentclass[conference]{IEEEtran}
\IEEEoverridecommandlockouts
\usepackage{amsmath,amssymb,amsfonts}
\usepackage{multicol}
\usepackage{algorithm}
\usepackage{algpseudocode}

\usepackage{graphicx}
\usepackage{textcomp}
\usepackage{xcolor}
\usepackage[caption=false, font=footnotesize]{subfig}
\usepackage{makecell}
\usepackage[super]{nth}

\usepackage{amsthm}
\theoremstyle{definition}

\RequirePackage{filecontents}
\usepackage[style=ieee]{biblatex}
\bibliography{reference}
\usepackage{comment}

\begin{document}

\title{Federated Learning for Sparse Principal Component Analysis\\
\thanks{This work was supported by the National Science and Technology Council, Taiwan, under grant 112-2634-F-001-001-MBK.}

}

\author{\IEEEauthorblockN{Sin Cheng Ciou\IEEEauthorrefmark{1}, Pin Jui Chen\IEEEauthorrefmark{2}, Elvin Y. Tseng\IEEEauthorrefmark{3} and Yuh-Jye Lee\IEEEauthorrefmark{4}}
\IEEEauthorblockA{\IEEEauthorrefmark{1}Dept. of Mathematics\\National Tsing Hua University, Hsinchu, Taiwan\\https://orcid.org/0009-0000-7372-1069}
\IEEEauthorblockA{\IEEEauthorrefmark{2}Dept. of Applied Mathematics\\National Yang Ming Chiao Tung University, Hsinchu, Taiwan\\Email: willie555577.sc11@nycu.edu.tw}
\IEEEauthorblockA{\IEEEauthorrefmark{3}Research Center for Information Technology Innovation\\Academia Sinica, Taipei, Taiwan\\https://orcid.org/0009-0009-2740-5829}
\IEEEauthorblockA{\IEEEauthorrefmark{4}Research Center for Information Technology Innovation\\Academia Sinica, Taipei, Taiwan\\Email: yuh-jye@citi.sinica.edu.tw}}

\markboth{IEEE TRANSACTIONS ON KNOWLEDGE AND DATA ENGINEERING, VOL.~60, NO.~12, DECEMBER~2021
}{Roberg \MakeLowercase{\textit{et al.}}: Federated Learning for Sparse Principal Component Analysis }

\maketitle
\begin{abstract}
In the rapidly evolving realm of machine learning, algorithm effectiveness often faces limitations due to data quality and availability. Traditional approaches grapple with data sharing due to legal and privacy concerns. The federated learning framework addresses this challenge. Federated learning is a decentralized approach where model training occurs on client sides, preserving privacy by keeping data localized. Instead of sending raw data to a central server, only model updates are exchanged, enhancing data security. We apply this framework to Sparse Principal Component Analysis (SPCA) in this work. SPCA aims to attain sparse component loadings while maximizing data variance for improved interpretability. {Beside the $\ell_1$ norm regularization term in conventional SPCA, we add a smoothing function to facilitate gradient-based optimization methods. Moreover, in order to improve computational efficiency, we introduce a least squares approximation to original SPCA. This enables analytic solutions on the optimization processes, leading to substantial computational improvements.} Within the federated framework, we formulate SPCA as a consensus optimization problem, which can be solved using the Alternating Direction Method of Multipliers (ADMM). Our extensive experiments involve both IID and non-IID random features across various data owners. Results on synthetic and public datasets affirm the efficacy of our federated SPCA approach.
\end{abstract}

\begin{IEEEkeywords}
ADMM, Consensus Learning, Federated Learning, Sparse Principal Component Analysis
\end{IEEEkeywords}

\section{Introduction}
In the rapidly evolving field of machine learning, the effectiveness of algorithms often hinges on the quantity, quality and accessibility of data. Unfortunately, due to legal and privacy concerns, data sharing for the purpose of enhancing machine learning is typically prohibited. To address this challenge, the Federated Learning (FL) framework was introduced \cite{mcmahan2017communication}. Federated learning represents a decentralized approach, where model training takes place on client-side devices, ensuring privacy by keeping data localized. Instead of transmitting raw data to a central server, only model updates are exchanged, thereby bolstering data security. This constitutes a distributed learning process.

In general, many machine learning tasks can be framed as optimization problems, and distributed optimization techniques have been well-developed. It is inherent in the nature of the federated learning framework to have applications across various machine learning algorithms, including linear and nonlinear support vector machines, as well as deep learning. In this context, we will introduce two innovative approaches: Federated Smoothing Sparse Principal Component Analysis (FSSPCA) and Federated Approximation Sparse Principal Component Analysis (FASPCA). These methodologies have been specifically crafted to tackle the challenge of Sparse Principal Component Analysis (SPCA) within a distributed framework.

Principal Component Analysis (PCA), introduced by Pearson \cite{pearson1901liii} in 1901, has been extensively used in various applications, from dimension reduction to anomaly detection \cite{lee2012anomaly}. The goal of PCA is to find an orthogonal basis that maximizes the variance of the original data when projected onto this basis. Formatting this as an optimization problem, we have
\begin{align*}
    \max_{\textbf{w} \in \mathbb{R}^{d \times r}} & \quad \| \textbf{A}\textbf{w} \|_{F}^{2}\\
    s.t. & \quad \textbf{w}^{\top}\textbf{w} = \textbf{I},
\end{align*}
where $\textbf{A}\in \mathbb{R}^{n \times d}$ is a mean-centered dataset with $n$ instances and $d$ features and $\|\cdot \|_{F}$ represents the Frobenius norm of a matrix. The matrix $\textbf{w} \in \mathbb{R}^{d \times r}$ consists of $r$ principal components, which are the eigenvectors corresponding to the $r$ largest eigenvalues of the covariance matrix of \textbf{A}. Alternatively, PCA can be reformulated into a minimization problem:
\begin{align*}
    \min_{\textbf{w} \in \mathbb{R}^{d \times r}} & \quad \| \textbf{A} - \textbf{A}\textbf{w}\textbf{w}^{\top} \|_{F}^{2}\\
    s.t. & \quad \textbf{w}^{\top}\textbf{w} = \textbf{I}.
\end{align*}
This reformulation can be interpreted as minimization of the \textit{reconstruction error}, which aims to represent data in fewer dimensions, while still approximating the original data as closely as possible. 

However, traditional PCA often yields dense component loadings, making the results difficult to interpret. To address this issue, Sparse PCA (SPCA) was proposed \cite{zou2006sparse}, aiming to find loadings that are more interpretable due to their sparsity.


In addition to its conventional applications, PCA is also harnessed in FL frameworks to enable collaborative computation without data centralization \cite{grammenos2020federated}. SPCA has yet to be fully adapted and studied within this emerging framework, suggesting a novel direction for advancement. In this work, we focus on a distributed setting in which various data owners aim to jointly train a model without sharing their own data with each other. Suppose that the objective function $f(\textbf{w})$ can be decomposed into $K$ parts as $f(\textbf{w}) = \sum_{i=1}^{K} f_{i}(\textbf{w})$, where $K$ is the number of data owners and each $f_{i}$ is a local objective function, which involves only the data of the $i$th owner. The minimization of $f(\textbf{w})$ can be reformulated as the following \textit{global variable consensus optimization} problem \cite{boyd2011distributed}:
\begin{align*}
    \min_{\textbf{w}_{1},\cdots,\textbf{w}_{K},\textbf{z}} & \quad \sum_{i=1}^{K} f_{i}(\textbf{w}_{i})\\
    s.t. & \quad \textbf{w}_{i}=\textbf{z}, \forall i,
\end{align*}
where $\textbf{z}$ is the common global variable and each $\textbf{w}_{i}$ is the local variable for each data owner. In the framework of FL, the alternating direction method of multipliers (ADMM) algorithm, a widely adopted approach in such context, can be employed.

The ADMM iterative algorithm within the framework of FL consists of two parts: the master's and the workers'. Suppose that there exists a central server, called the master, responsible for updating the global variable $\textbf{z}$ and integrating all local variables $\textbf{w}_{i}$. Each data owner employs a device known as the worker, tasked with updating the local variable $\textbf{w}_{i}$ by minimizing the objective $f_{i}$ using its own data. Following the update of $\textbf{w}_{i}$, each worker sends the locally updated variable $\textbf{w}_{i}$ to the master. Consequently, the master integrates $\textbf{w}_{i},\forall i$, to update $\textbf{z}$, and then distributes the globally updated parameter $\textbf{z}$ to all workers. The goal of this iterative process is to compute the sparse loadings within the FL framework.

\section{RELATED WORK}
\subsection{Sparse Principal Component Analysis}
In recent years, numerous approaches for SPCA have been introduced \cite{zou2018selective}. Inspired by lasso regression, Jolliffe et al. \cite{jolliffe2003modified} proposed SCOTLASS, which added an $\ell_{1}$-norm term to the original PCA problem. 
Unlike SCOTLASS, Zou et al. \cite{zou2006sparse} added an $\ell_{1}$-norm term or an elastic net term by
\begin{equation}
    \begin{aligned}
        \label{eq:SPCA_zou}
        \min_{\textbf{w}, \textbf{v} \in \mathbb{R}^{d \times r}} & \quad \| \textbf{A} - \textbf{A}\textbf{w}\textbf{v}^{\top} \|_{F}^{2} + \sum_{i = 1}^{r} \left( \lambda \| v_{i} \|_{2}^{2} +  \gamma_{i} \| v_{i} \|_{1} \right) \\
        s.t. & \quad \textbf{w}^{\top}\textbf{w} = \textbf{I},
    \end{aligned}
\end{equation}
where $\textbf{A}\in\mathbb{R}^{n\times d}$ is a mean-centered dataset of $n$ instances with $d$ features, $r$ is the number of desired loadings, $v_{i}$ is the $i$th column of $\textbf{v}$, and $\lambda, \gamma_{i}>0$ are trade-off parameters. 
They solved this problem by iterating the sub-problems of $\textbf{w}$ and $\textbf{v}$ alternately. 
One advantage of this approach is that it decomposes the orthogonality constraint and the non-smooth $\ell_{1}$-norm term into two sub-problems. Thus, each sub-problem is easier to solve than the original problem. Some methods tackle SPCA by relying on power methods. Journee et al. \cite{journee2010generalized} proposed the generalized power method (Gpower) to extract a single sparse dominant principal component or more components at once with the $\ell_{0}$-norm or the $\ell_{1}$-norm. Ge et al. \cite{ge2018minimax} proposed distributed privacy-preserving SPCA to compute SPCA in a distributed optimization framework by the power iteration with differential privacy.




\subsection{ADMM}
The Alternating Direction Method of Multipliers (ADMM) is an optimization method commonly used in federated learning. It was first proposed in the 1970s by Glowinski \cite{glowinski1975approximation} and Gabay et al. \cite{gabay1976dual}. In 2011, Boyd et al. \cite{boyd2011distributed} proposed a detailed review of this algorithm, making it widely adopted in decentralized machine learning \cite{silva2019federated,lee2019CRSVM,he2018cola}. Furthermore, by combining cryptographic tools such as multiparty
computation \cite{damgaard2012multiparty,cramer2001multiparty}, and zero-knowledge proof \cite{garay2003strengthening}, the ADMM algorithm can protect data privacy when applied to federated learning models \cite{zheng2019helen,smith2018cocoa}

Some studies employed the ADMM algorithm on sparse PCA. Ma and Shiqian \cite{ma2013alternating} solved the DSPCA by the ADMM algorithm. Vu et al. \cite{vu2013fantope} viewed the SPCA problem as an optimization problem on the Fantope and then used the ADMM algorithm to solve it. Instead of the convex penalty term $\ell_{1}$-norm, Hajinezhad and Hong \cite{hajinezhad2015nonconvex} used non-convex penalties to obtain sparse loading by the non-convex ADMM algorithm. Tan et al. \cite{tan2019learning} considered the SPCA problem as an optimization problem on the Stiefel manifold and then used the ADMM algorithm to solve it. Despite the widespread use of ADMM in SPCA, these studies have not yet ventured into its application within the context of federated learning.

\subsection{Smoothing function}
When dealing with optimization problems, deriving closed-form solutions is often not practical for the majority of cases. In such scenarios, it is a common approach to use gradient-based methods such as Newton's method, stochastic gradient descent, and Adam method \cite{kingma2014adam}. Nevertheless, if the problem lacks differentiability, it remains unfeasible to employ these gradient-based methods. For example, because of the $\ell_{1}$-norm, problem \eqref{eq:SPCA_zou} is not differentiable at 0. To address this issue, a common approach is to employ a smoothing function as a replacement for the non-differentiable part \cite{hebiri2011smooth,saheya2018numerical,saheya2019neural}. This technique is also widely adopted in the field of machine learning and deep learning. For instance, the sigmoid function is often used to substitute the rectified linear unit \cite{lee2001ssvm,lee2005spl}. However, to our knowledge, using smoothing functions to replace the $\ell_{1}$-norm in the SPCA has not been explored. In our proposed method, we employ a smoothing function to substitute the $\ell_{1}$-norm, thus enabling the utilization of gradient-based methods to the SPCA problem.

\section{Proposed model}

\subsection{Federated Smoothing Sparse Principal Component Analysis}
Let $\textbf{A} = [\textbf{A}^{1},\cdots,\textbf{A}^{K}]^{\top} \in \mathbb{R}^{n\times d}$ be a mean-centered dataset comprising $n$ instances with $d$ features, stored separately in the $K$ distinct devices. Each worker can only access its own local data, $\textbf{A}^{i} \in \mathbb{R}^{n_{i}\times d}$, where $n_{i}$ represents the number of instances in the dataset for each worker. Throughout the computation, the workers can only share model parameters but not data by uploading the locally updated parameter to the central server (master). In this work, we aim to solve the following optimization problem to find sparse loadings of the dataset $\textbf{A}$:
\begin{equation}
    \begin{aligned}
        \label{eq:SPCA_Prime_Muti}
        \min_{ \textbf{w} \in \mathbb{R}^{d \times r}} & \quad \| \textbf{A} - \textbf{A}\textbf{w}\textbf{w}^{\top} \|_{F}^{2} + \lambda\|\textbf{w}\|_{1} \\
        s.t. & \quad \textbf{w}^{\top}\textbf{w} = \textbf{I},
    \end{aligned}
\end{equation}
where $r$ is the number of desired loadings, $\lambda \geq 0$ is a trade-off parameter, $\textbf{A}\textbf{w}\textbf{w}^{\top}$ is the reconstructed data using $\textbf{w}$, and $\|\cdot\|_{1}$ represents the $\ell_{1}$-norm for matrix as$\|\textbf{w}\|_{1} = \sum_{i,j}|w_{ij}|$. 

The first term of \eqref{eq:SPCA_Prime_Muti} is to minimize the reconstruction error, and the second term is to pursue the sparse representation through the $\ell_{1}$-norm, with the constraint of maintaining the orthonormality of the loading. However, the orthonormality constraint is challenging. Following Tan et al. \cite{tan2019learning}, we modify it by restricting $\textbf{w}$ to the Stiefel manifold,
\begin{align*}
    \mathcal{M}_{r} = \{ \textbf{w} \in \mathbb{R}^{d \times r} | \textbf{w}^{\top}\textbf{w} = \textbf{I} \},
\end{align*}
which is the set of $d$-by-$r$ orthonormal matrices to simplify the problem. Subsequently, we can reformulate \eqref{eq:SPCA_Prime_Muti} as follows
\begin{align}
    \label{eq:SPCA_Simple_Stiefel}
    \min_{ \textbf{w} \in \mathcal{M}_{r} } \| \textbf{A} - \textbf{A}\textbf{w}\textbf{w}^{\top} \|_{F}^{2} + \lambda\|\textbf{w}\|_{1}.
\end{align}

Though, the orthonormality constraint of \eqref{eq:SPCA_Simple_Stiefel} and the non-smooth $\| \textbf{w} \|_{1}$ term still remain challenging in solving the problem. Moreover, since each worker cannot share their own dataset $ \textbf{A}^{i} $ with  each other, it suffices to reformulate \eqref{eq:SPCA_Simple_Stiefel} in a distributed form. Note that $\| \textbf{A} - \textbf{A}\textbf{w}\textbf{w}^{\top} \|_{F}^{2}$ can be decompose into $K$ parts by
\begin{align}
    \| \textbf{A} - \textbf{A}\textbf{w}\textbf{w}^{\top} \|_{F}^{2} = \sum_{i=1}^{K}\| \textbf{A}^{i} - \textbf{A}^{i}\textbf{w}\textbf{w}^{\top} \|_{F}^{2}.
\end{align}
Hence, we can reformulate \eqref{eq:SPCA_Simple_Stiefel} as a consensus optimization problem,
\begin{equation}
    \begin{aligned}       
        \label{eq:Tan_opt}
        \min_{ \textbf{w}_{1},\cdots,\textbf{w}_{K} \in \mathcal{M}_{r} } & \quad \sum_{i=1}^{K} \| \textbf{A}^{i} - \textbf{A}^{i}\textbf{w}_{i}\textbf{w}_{i}^{\top} \|_{F}^{2} + \lambda\|\textbf{z}\|_{1} \\
        s.t. & \quad \textbf{w}_{i} = \textbf{z}, \forall  i,
    \end{aligned}
\end{equation}
where $\textbf{w}_{1},\cdots,\textbf{w}_{K}$ are parameters for each worker and $\textbf{z} \in \mathbb{R}^{d \times r}$ is the consensus parameter. With this formulation, we can adopt the algorithm in Tan et al. \cite{tan2019learning}.

However, we observe that the algorithm struggles to converge when applied to the federated learning problem. 
As shown in Fig. \ref{fig:FSSPCA_WDBC_counter_L_0}, the objective function exhibits significant fluctuations during the iterative process. 
To solve this problem, we introduce the Federated Smoothing SPCA (FSSPCA) in the following.

\begin{figure}[htbp]
    \centering
    \includegraphics[width=0.8\linewidth]{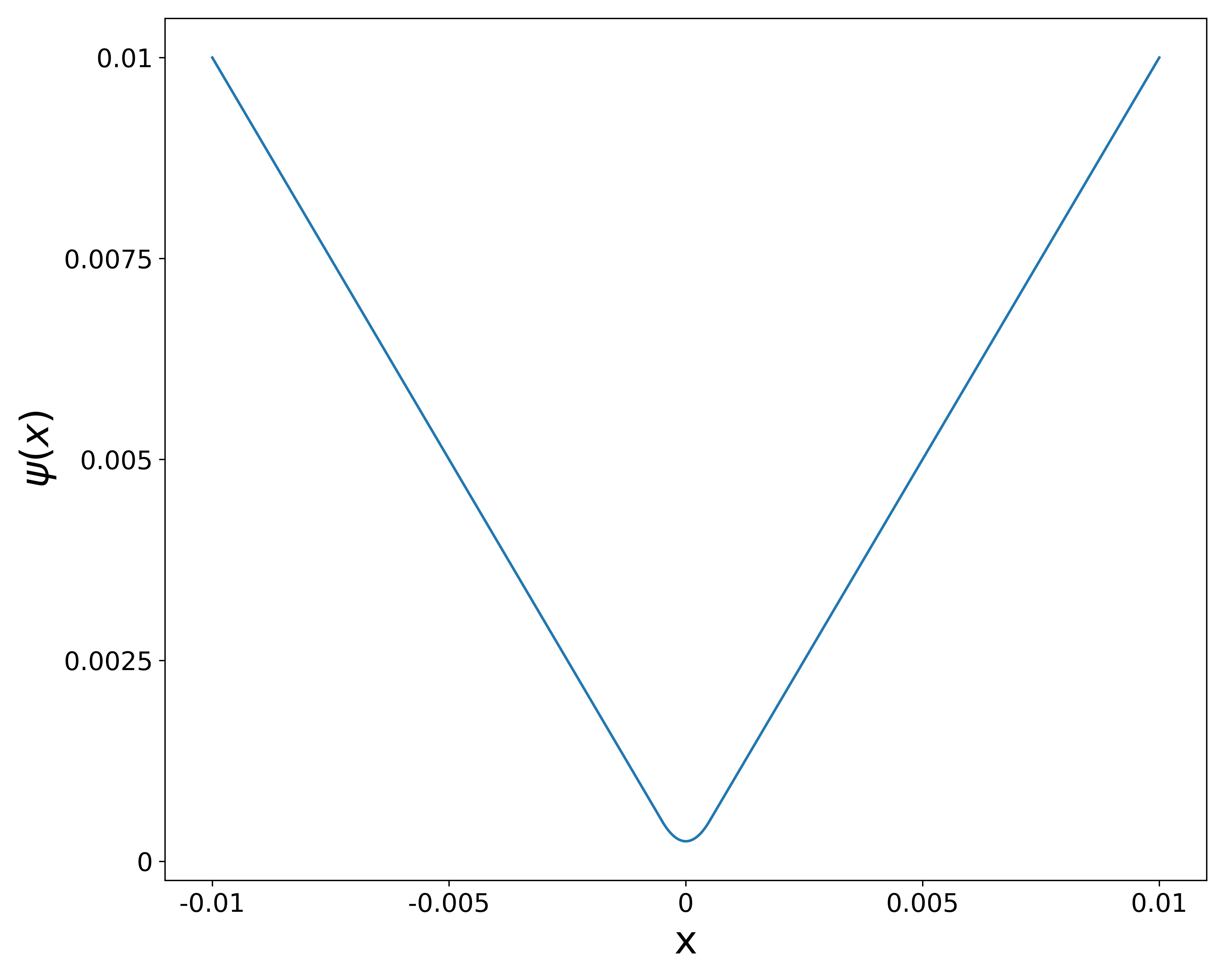}
    \caption{The diagram of the $\ell_{1}$-norm smoothing function with $\mu=10^{-3}$.}
    \label{fig:l1_smoothing_function}
\end{figure}

Inspired by Saheya et al. \cite{saheya2018numerical}, we add an $\ell_{1}$-norm smoothing function $r(\cdot)$ to problem \eqref{eq:Tan_opt}
\begin{equation}
    \begin{aligned}
        \label{eq:FSSPCA_Prime}
        \min_{ \textbf{w}_{1},\cdots,\textbf{w}_{K} \in \mathcal{M}_{r} } & \quad \sum_{i=1}^{K} \left( \| \textbf{A}^{i} - \textbf{A}^{i}\textbf{w}_{i}\textbf{w}_{i}^{\top} \|_{F}^{2} + \lambda_{1}r(\textbf{w}_{i}) \right)+ \lambda_{2}\|\textbf{z}\|_{1} \\
        s.t. & \quad \textbf{w}_{i} = \textbf{z}, \forall  i,
    \end{aligned}
\end{equation}
where $\lambda_{1}, \lambda_{2} \geq 0$ are the trade-off parameters. Given a real matrix $\textbf{X}$, the $\ell_{1}$-norm of $\textbf{X}$ is the sum of all absolute values of entries in $\textbf{X}$. Therefore, it is reasonable to construct the smoothing function $r(\cdot)$ using an absolute value smoothing function $\psi(\cdot)$. Following Saheya et al. \cite{saheya2018numerical,saheya2019neural}, we use $r(\textbf{X}) = \sum_{i,j}\psi(x_{i,j})$ due to its accuracy and computational efficiency, as described by the following equation:
\begin{align}
    \psi(x_{i,j}) = \left\{ \begin{array}{rcl}
    x_{i,j} & \mbox{if} & x_{i,j} \geq \frac{\mu}{2} \\
    \frac{x_{i,j}^{2}}{\mu} + \frac{\mu}{4} & \mbox{if} & \frac{-\mu}{2} < x_{i,j} < \frac{\mu}{2} \\
    -x_{i,j} & \mbox{if} & x_{i,j} \leq \frac{-\mu}{2}
\end{array}\right. ,
\end{align}
where $\mu$ is a similarity parameter for $\psi(x_{i,j})$ such that $ \lim_{\mu \to 0} \psi(x_{i,j}) \to | x_{i,j} | $. In our experiments, we set $\mu$ to $10^{-3}$. As depicted in Fig. \ref{fig:l1_smoothing_function}, when $x_{i,j}$ is closed to zero, the smoothing function $\psi(x_{i,j})$ closely approximates the absolute value of $x_{i,j}$, providing a smoothed representation.

We can observe that the consensus optimization formulation \eqref{eq:FSSPCA_Prime} separates the orthonormality constraint and the non-smooth term into two sub-problems, making each more accessible. Subsequently, following Boyd et al. \cite{boyd2011distributed}, we use the ADMM algorithm to solve this problem. By introducing a dual variable $\textbf{u}_{i} \in \mathbb{R}^{d \times r}$ for the equality constraint, we obtain an augmented Lagrangian:
\begin{equation}
    \begin{aligned}
        \mathcal{L}_{\rho}^{S} ( \{ \textbf{w}_{i}, \textbf{u}_{i} \}_{i=1}^{K}, \textbf{z} ) &= \sum_{i=1}^{K} \left( \vphantom{\frac{\rho}{2}} \| \textbf{A}^{i} - \textbf{A}^{i}\textbf{w}_{i}\textbf{w}_{i}^{\top} \|_{F}^{2} + \lambda_{1}r(\textbf{w}_{i}) \right.\\
        & \left. \vphantom{\| \textbf{A}^{i} - \textbf{A}^{i}\textbf{w}_{i}\textbf{w}_{i}^{\top} \|_{F}^{2}} + \langle \textbf{u}_{i}, \textbf{w}_{i} - \textbf{z} \rangle + \frac{\rho}{2} \| \textbf{w}_{i} - \textbf{z} \|_{F}^{2} \right) + \lambda_{2}\|\textbf{z}\|_{1},
    \end{aligned}
\end{equation}
where $\rho >0$ is a penalty parameter. The general ADMM algorithm for FSSPCA is outlined in Algorithm \ref{alg:FSSPCASeudoCode}. The algorithm minimizes $\mathcal{L}_{\rho}^{S}(\cdot)$, and employs the function $\text{qf}(\cdot)$, which denotes the Q factor in the QR decomposition, to ensure the orthogonality of the loadings in the last step.

\begin{algorithm}
\caption{ADMM Algorithm for FSSPCA}
\label{alg:FSSPCASeudoCode}
\begin{algorithmic}[1]
\Require
    Datasets $ \textbf{A}^{1},\cdots,\textbf{A}^{K}$, $\lambda_{1}$, $\lambda_{2}$, $\rho$
\Ensure The sparse loadings $\textbf{z}$
\State Initialize $\textbf{w}_{i}^{0}\in\mathbb{R}^{d \times r}$ by a random orthonormal matrix, $\forall i$,  $\textbf{u}_{i}^{0}, \textbf{z}^{0} = \textbf{0}, \ \forall i$, and $t = 0$
\While{stopping criteria are not satisfied}

\noindent\texttt{Workers:} $// \ In\ parallel$ \Comment{local primal update}
    \For{$i \in\{ 1,\dots, K\}$}
        \State $ \textbf{w}_{i}^{t+1} \gets \operatorname*{argmin}_{ \textbf{w}_{i} \in \mathcal{M}_{r}} \mathcal{L}_{\rho}^{S}( \{ \textbf{w}_{i}, \textbf{u}_{i}^{t} \}_{i=1}^{K}, \textbf{z}^{t} )$
        \State Send $ \textbf{w}_{i}^{t+1} $ to the master
    \EndFor

\noindent\texttt{Master:} \Comment{consensus primal update}
    \State $ \textbf{z}^{t+1} \gets \operatorname*{argmin}_{\textbf{z}} \mathcal{L}_{\rho}^{S}( \{ \textbf{w}_{i}^{t+1}, \textbf{u}_{i}^{t} \}_{i=1}^{K}, \textbf{z} ) $
    \State Send $ \textbf{z}^{t+1} $ to the all workers

\noindent\texttt{Workers:} $// \ In\ parallel$ \Comment{local dual update}
    \For{$i \in\{ 1,\dots, K\}$}
        \State $ \textbf{u}_{i}^{t+1} \gets \textbf{u}_{i}^{t} + \rho(\textbf{w}_{i}^{t+1} - \textbf{z}^{t+1})$
    \EndFor
    \State $t \gets t + 1$
\EndWhile
\State $\textbf{z} \gets \text{qf}(\textbf{z}^{t})$
\end{algorithmic}
\end{algorithm}

Next, let us introduce how to compute $\textbf{w}_{i}$ and $\textbf{z}$ in Algorithm \ref{alg:FSSPCASeudoCode}. For the $\textbf{z}$ update, given fixed $\{ \textbf{w}_{i}, \textbf{u}_{i} \}_{i=1}^{K}$, the optimization of $ \mathcal{L}_{\rho}^{S}( \{ \textbf{w}_{i}, \textbf{u}_{i} \}_{i=1}^{K}, \textbf{z} ) $ over $\textbf{z}$ is equivalent to

\begin{align}
    \label{eq:ZupdatingScale}
    \min_{\textbf{z}} \quad \lambda_2\|\textbf{z}\|_{1} + \frac{K\rho}{2} \| \textbf{z} - ( \bar{\textbf{w}} + \frac{1}{\rho} \bar{\textbf{u}} ) \|_{2}^{2},
\end{align}
which has a closed-form solution $ \textbf{z}^{\star}$. In which $ \bar{\textbf{w}} = \sum_{i=1}^{K} \textbf{w}_{i}/K $ and $ \bar{\textbf{u}} = \sum_{i=1}^{K} \textbf{u}_{i}/K $, by defining $ \textbf{v} := \bar{\textbf{w}} +\bar{\textbf{u}}/\rho$, the update for $ \textbf{z}^{\star} $ can be computed using the soft-thresholding operator
\begin{align}
    \label{eq:SoftThreshold}
    z_{ij}^{\star} := \max( | v_{ij} | - \frac{\lambda_2}{K\rho}, 0)sign(v_{ij}),
\end{align}
where $z_{ij}^{\star}, v_{ij}$ are the entries of $\textbf{z}^{\star}, \textbf{v}$, respectively.

\begin{algorithm}
\caption{Master's Update for FSSPCA}
\label{alg:FSSPCA_z_update}
\begin{algorithmic}[1]
\Require
    $ \{ \textbf{w}_{i}^{t+1}, \textbf{u}_{i}^{t} \}_{i=1}^{K}$, $\lambda_{2}$, $\rho $
\Ensure $\textbf{z}^{t+1}$
\State $\textbf{z}^{t+1} \gets \dfrac{1}{K}\sum_{i=1}^{K} (\textbf{w}_{i}^{t+1} + \dfrac{1}{\rho}\textbf{u}_{i}^{t} )$
\State $ \textbf{z}_{ij}^{t+1} \gets \max( | \textbf{z}_{ij}^{t+1} | - \frac{\lambda_{2}}{K\rho}, 0)sign(\textbf{z}_{ij}^{t+1})$
\end{algorithmic}
\end{algorithm}

In the next subsection, we introduce how to update $\textbf{w}_{i}$ (local primal update) in Algorithm \ref{alg:FSSPCASeudoCode}.

\subsection{Line-search method on the Stiefel Manifold}
For each worker, with fixed $\textbf{u}_{i}$ and $\textbf{z}$, the optimization of $ \mathcal{L}_{\rho}^{S}( \{ \textbf{w}_{i}, \textbf{u}_{i} \}_{i=1}^{K}, \textbf{z} ) $ over $\textbf{w}_{i}$ is given by
\begin{equation}
    \begin{aligned}
        \label{eq:FSSPCA_w_updating}
        \min_{\textbf{w}_{i} \in \mathcal{M}_{r}} \quad & \| \textbf{A}^{i} - \textbf{A}^{i}\textbf{w}_{i}\textbf{w}_{i}^{\top} \|_{F}^{2} + \lambda_{1}r(\textbf{w}_{i})\\
        & + \langle \textbf{u}_{i}, \textbf{w}_{i}  \rangle + \frac{\rho}{2} \| \textbf{w}_{i} - \textbf{z} \|_{F}^{2}.
    \end{aligned}
\end{equation}
Following Tan et al. \cite{tan2019learning}, \eqref{eq:FSSPCA_w_updating} can be solved using the line-search method on the Stiefel Manifold. 

The primary objective of the line search method is to find a descent direction in which our objective function decreases. A direct approach is to take the opposite direction of the gradient as the descent direction. However, since the objective function \eqref{eq:FSSPCA_w_updating} is restricted to the Stiefel manifold, its gradient cannot be directly computed through partial differentiation. 

To address optimization on the Stiefel manifold $\mathcal{M}_{r}$, we introduce some geometries over the Stiefel manifold $\mathcal{M}_{r}$. First, the tangent space to $\mathcal{M}_{r}$ at $\textbf{w} \in \mathcal{M}_{r}$, denoted by $T_{\textbf{w}}\mathcal{M}_{r}$, is defined as
\begin{align*}
    T_{\textbf{w}}\mathcal{M}_{r} = \{ \textbf{z} \in \mathbb{R}^{d\times r} | \textbf{z}^{\top}\textbf{w} + \textbf{w}^{\top}\textbf{z} = \textbf{0} \}.
\end{align*}
Given a smooth function $f$ on Euclidean space, let $\bar{f}$ denote the restriction of $f$ to the Stiefel manifold. Define sym($\textbf{A}$) = $\frac{1}{2}( \textbf{A} + \textbf{A}^{\top} )$, for any square matrix $\textbf{A}$. Suppose we have the gradient $\nabla f(\textbf{w})$ of $f$ at $\textbf{w} \in \mathcal{M}_{r}$. Then the gradient of $\bar{f}(\textbf{W)}$, denoted by grad$f(\textbf{w})$, is equal to the orthogonal projection of $\nabla f(\textbf{w})$ onto $T_{\textbf{w}}\mathcal{M}_{r}$, that is,
\begin{align}
    \text{grad}f(\textbf{w}) = P_{T_{\textbf{w}}\mathcal{M}_{r}}(\nabla f(\textbf{w})),
\end{align}
where $ P_{T_{\textbf{w}}\mathcal{M}_{r}} $ is denoted the orthogonal projection function onto $T_{\textbf{w}}\mathcal{M}_{r}$ at $\textbf{w} \in \mathcal{M}_{r}$ by 
\begin{align}
    P_{T_{\textbf{w}}\mathcal{M}_{r}}(\textbf{X}) = \textbf{X} - \textbf{w}\text{ sym}( \textbf{w}^{\top} \textbf{X} ),
\end{align}
for any $\textbf{X}\in \mathbb{R}^{d \times r}$.

After obtaining the gradient of the objective function, which is restricted to the Stiefel manifold, we can proceed to construct the descent direction $\boldsymbol{\zeta} = -\text{grad}f(\textbf{w})$. Additionally, the use of conjugate gradient \cite{tan2019learning,absil2009optimization} can improve the convergence speed at workers' stage. However, the increase of local updates might escalate the divergence between the resultant local models at the workers' stage before they are averaged \cite{kairouz2019advances}. Therefore, for stability, we still use the opposite gradient as the descent direction.

With step size $\theta$, we can move $\textbf{w}$ along the descent direction $\boldsymbol{\zeta}$ as $\textbf{w}_{next} = \textbf{w} + \theta\boldsymbol{\zeta}$. Unlike the optimization in the Euclidean space, the update process $\textbf{w}$ requires ensuring that $\textbf{w}_{next}$ lies within the confines of the Stiefel manifold. To achieve this, we employ the \textit{Retraction} mapping to project it onto the manifold by
\begin{align}
    R_{\textbf{w}}(\theta\boldsymbol{\zeta}) = \text{qf}(\textbf{w} + \theta\boldsymbol{\zeta}),
\end{align}
where $\text{qf}(\cdot)$ denotes the Q factor in the QR decomposition ensuring the orthonormality.

Finally, following Ring and Wirth \cite{ring2012optimization}, we choose the step size $\theta$ satisfying the strong Wolfe conditions
\begin{align}
    \label{eq:Wolfe1}
    f(R_{\textbf{w}}(\theta\boldsymbol{\zeta})) \leq f(\textbf{w}) + c_{1}\theta\langle \text{grad}f(\textbf{w}), \boldsymbol{\zeta} \rangle,\\
    \label{eq:Wolfe2}
    | \langle \text{grad}f(R_{\textbf{w}}), \boldsymbol{\zeta} \rangle | \leq -c_{2} \langle \text{grad}f(\textbf{w}), \boldsymbol{\zeta} \rangle,
\end{align}
where $0 < c_{1} < c_{2} < 1$. Alternatively, in pursuit of simplicity, one can follow Absil et al. \cite{absil2009optimization} to choose a step size that fulfills the Armijo condition \eqref{eq:Wolfe1}.

\begin{algorithm}
\caption{Worker's Update for FSSPCA on the Stiefel Manifold}
\label{alg:FSSPCA_worker_updating}
\begin{algorithmic}[1]
\Require
    Objective function $f : \mathcal{M}_{r} \rightarrow \mathbb{R}$
\Ensure
   The minimum $\textbf{w}$ of the objective function $f$
\State Initialize $\textbf{w}^{0}\in\mathbb{R}^{d \times r}$ by a random orthogonal matrix, and $t = 0$
\While{stopping criteria are not satisfied}
    \State Compute $ \operatorname{grad}f(\textbf{w}^{t+1}) \gets P_{T_{\textbf{w}^{t}}\mathcal{M}_{r} }(\triangledown f(\textbf{w}^{t})) $
    \State Compute $ \boldsymbol{\zeta}^{t+1} \gets -\operatorname{grad}f(\textbf{w}^{t+1}) $
    \State Choose a step size $ \theta^{t+1} \in \mathbb{R}$, where $ \theta^{t+1} $ satisfying the strong Wolfe (Armijo) conditions
    \State Set $ \textbf{w}^{t+1} \gets R_{\textbf{w}^{t}}(\theta^{t+1}\boldsymbol{\zeta}^{t+1}) $
    \State Set $t \gets t+1$
\EndWhile
\State $\textbf{w} \gets \textbf{w}^{t}$
\end{algorithmic}
\end{algorithm}

\subsection{Deflation method for FSSPCA}
FSSPCA can compute multiple loadings simultaneously. However, for the sake of completeness, we further introduce the method of computing additional $r_2$ loadings, considering the existence of $r_1$ loadings, without the necessity of recomputing all of them. Drawing upon the deflation technique for SPCA proposed by Mackey \cite{mackey2008deflation}, we outline the deflation scheme in Algorithm \ref{alg:SmoFSPCA_Defla}.

\begin{algorithm}
\caption{Deflation Method for FSSPCA}
\label{alg:SmoFSPCA_Defla}
\begin{algorithmic}[1]
\Require
    Datasets $ \textbf{A}^{1},\cdots,\textbf{A}^{K}$, the number of iterations of the deflation method $T$, the number of loadings $\{r_{j}\}_{j=1}^{T}$ we want to compute an each iterations, the parameters $\{\lambda_{1j},\lambda_{2j},\rho_{j}\}_{j=1}^{T}$, an empty array $\textbf{z}$
\Ensure Sparse loadings $\textbf{z}$
\State Initialize $ \textbf{G}_{0} = \textbf{I}\in \mathbb{R}^{d \times d}$
\For{$j = 1 : T$}
    \State $ \textbf{z}_{j} \gets \operatorname*{argmin}_{\textbf{z}} (\sum_{i=1}^{K} \| \textbf{A}^{i} - \textbf{A}^{i}\textbf{w}_{i} \textbf{w}_{i}^{\top} \|_{2}^{2}  + \lambda_{1j}r(\textbf{w}_{i})) + \lambda_{2j}\|\textbf{z}\|_{1}$
    \State $ \textbf{z}_{j} \gets \textbf{G}_{j-1} \textbf{z}_{j}$
    \For{$i = 1 : K$}
        \State $ \textbf{A}^{i} \gets \textbf{A}^{i} ( \textbf{I} - \textbf{z}_{j}\textbf{z}_{j}^{\top} )$
    \EndFor
    \State $ \textbf{G}_{j} \gets \textbf{G}_{j-1}( \textbf{I} - \textbf{z}_{j}\textbf{z}_{j}^{\top} )$
    \State $ \textbf{z}_{j} \gets \text{qf}(\textbf{z}_{j})$
    \State $\textbf{z} \gets [\textbf{z}, \textbf{z}_{j}]$
\EndFor
\end{algorithmic}
\end{algorithm}

Suppose we have $r_1$ loadings $\{ \textbf{z}_{i} \}_{i=1}^{r-1}$ and intend to compute additional $r_2$ loadings, the problem is formulated as 
\begin{equation}
    \begin{aligned}
        \label{eq:FSSPCA_Prime_Defla}
        \min_{ \textbf{w}_{1},\cdots,\textbf{w}_{K} \in \mathcal{M}_{r_2} } & \quad \sum_{i=1}^{K} \left( \| \textbf{A}^{i} - \textbf{A}^{i}\textbf{w}_{i}\textbf{w}_{i}^{\top} \|_{F}^{2} + \lambda_{1}r(\textbf{w}_{i}) \right)+ \lambda_{2}\|\textbf{z}\|_{1} \\
        s.t. & \quad \textbf{w}_{i}^{\top}\textbf{G}\textbf{w}_{i} = \textbf{I} \text{ and } \textbf{w}_{i} = \textbf{z},\ \forall  i,
    \end{aligned}
\end{equation}
where $\textbf{G} = \prod_{i=1}^{r_1} ( \textbf{I} - \textbf{z}_{i} \textbf{z}_{i}^{\top}) \in \mathbb{R}^{d \times d}$ lie in the orthogonal complement space of $\{ \textbf{z}_{i} \}_{i=1}^{r_1}$. 
Because of the orthogonal complement space $\textbf{G}$, the objective function \eqref{eq:FSSPCA_Prime_Defla} is actually restricted to the generalized Stiefel manifold $\mathcal{M}_{r_{2}}^{\textbf{G}}$,
\begin{align*}
    \mathcal{M}_{r_{2}}^{\textbf{G}} = \{ \textbf{w} \in \mathbb{R}^{d \times r_{2}} | \textbf{w}^{\top}\textbf{G}\textbf{w} = \textbf{I} \},
\end{align*}
which is the set of $d$-by-$r_{2}$ $\textbf{G}$-orthogonal matrices. By restricting to $\mathcal{M}_{r_{2}}^{\textbf{G}}$, we can ensure that the new loadings $\{ \textbf{z}_{i} \}_{i=r_{1}+1}^{r_{1}+r_{2}}$ must be orthogonal to the existing loadings.

Considering the difference between the Stiefel manifold $\mathcal{M}_{r}$ and the generalized Stiefel manifold $\mathcal{M}_{r_{2}}^{\textbf{G}}$, we extend Algorithm \ref{alg:FSSPCA_worker_updating} to fit the deflation scenario. First, the tangent space to $\mathcal{M}_{r_{2}}^{\textbf{G}}$ at $\textbf{w}  \in \mathcal{M}_{r_{2}}^{\textbf{G}}$, denoted by $T_{\textbf{w}}\mathcal{M}_{r_{2}}^{\textbf{G}}$, is defined as
\begin{align*}
    T_{\textbf{w}}\mathcal{M}_{r_{2}}^{\textbf{G}} = \{ \textbf{z} \in \mathbb{R}^{d\times r_{2}} | \textbf{z}^{\top}\textbf{G}\textbf{w} + \textbf{w}^{\top}\textbf{G}\textbf{z} = \textbf{0} \}.
\end{align*}
Then the orthogonal projection operator onto $T_{\textbf{w}}\mathcal{M}_{r_{2}}^{\textbf{G}}$, denoted by $ P_{T_{\textbf{w}}\mathcal{M}_{r_{2}}^{\textbf{G}}} $, is gien by
\begin{align}
    \label{eq:gen_stiefl_ten_pro}
    P_{T_{\textbf{w}}\mathcal{M}_{r_{2}}^{\textbf{G}}}(\textbf{z}) = \textbf{z} - \textbf{w}\text{ sym}( \textbf{w}^{\top} \textbf{G} \textbf{z} ).
\end{align}
Moreover, the retraction for the generalized Stiefel manifold is modified as
\begin{align}
    \label{eq:Retra_gener}
    R_{\textbf{w}}^{\textbf{G}}(\theta\boldsymbol{\zeta}) = \sqrt{\textbf{G}}^{-1} \text{qf} \left( \sqrt{\textbf{G}}( \textbf{w} +  \theta\boldsymbol{\zeta} ) \right).
\end{align}
Nevertheless, when computing \eqref{eq:Retra_gener}, we require $\sqrt{\textbf{G}}$ and $\sqrt{\textbf{G}}^{-1}$, which may be computationally expensive for large $d$. To address this problem, Sato and Aihara \cite{sato2019cholesky} proposed the Cholesky QR-based retraction as shown in Algorithm \ref{alg:Chol_Retraction}.

\begin{algorithm}
\caption{ Cholesky QR-based Retraction on the Generalized Stiefel manifold }
\label{alg:Chol_Retraction}
\begin{algorithmic}[1]
\Require
    A point $ \textbf{w} \in \mathcal{M}_{r_{2}}^{\textbf{G}} $, a step size $\theta$, and the descent direction $\boldsymbol{\zeta}$.
\Ensure
   $R_{\textbf{w}}^{\textbf{G}}(\theta\boldsymbol{\zeta})$.
\State $\textbf{z} \gets (\textbf{w} + \theta\boldsymbol{\zeta})^{\top} \textbf{G} (\textbf{w} + \theta\boldsymbol{\zeta}) $.
\State Compute the Cholesky factorization of $\textbf{z}$ as $\textbf{z} = \textbf{R}^{\top}\textbf{R}$.
\State $ R_{\textbf{w}}^{\textbf{G}}(\theta\boldsymbol{\zeta}) = (\textbf{w} + \theta\boldsymbol{\zeta})\textbf{R}^{-1}$.
\end{algorithmic}
\end{algorithm}

Thus, replacing the projection in Algorithm \ref{alg:FSSPCA_worker_updating} with \eqref{eq:gen_stiefl_ten_pro} and retraction with Algorithm \ref{alg:Chol_Retraction}, we manage to extend Algorithm \ref{alg:FSSPCA_worker_updating} from the Stiefel manifold to the generalized Stiefel manifold.

\subsection{Federated Approximation Sparse Principal Component Analysis}
In the previous subsection, we introduced how to tackle the conventional SPCA problem within the federated learning system. However, we observed that employing gradient-based methods to compute $r$ loadings simultaneously could incur significant computational costs. Regarding the scenarios that only a single or the first few loadings are desired, we propose an alternative approach using a least squares approximation to alleviate the computational burden. The problem is formulated as 
\begin{equation}
    \begin{aligned}
        \label{eq:SPCA_Prime}
        \min_{ \textbf{w} \in \mathbb{R}^{d \times 1}} & \quad \| \textbf{A} - \textbf{A}\textbf{w}\textbf{w}^{\top} \|_{2}^{2} + \lambda\|\textbf{w}\|_{1} \\
        s.t. & \quad \textbf{w}^{\top}\textbf{w} = 1,
    \end{aligned}
\end{equation}
where $\lambda \geq 0$ is a trade-off parameter, and $\textbf{w}\in \mathbb{R}^{d\times 1}$ is the desired loading.
Inspired by Lee et al. \cite{lee2012anomaly,papadimitriou2005streaming}, we propose the Federated Approximation SPCA (FASPCA) to solve this problem. Following Yang \cite{yang1995projection}, we utilize the projection approximation technique to transform \eqref{eq:SPCA_Prime} into the subsequent form, 
\begin{equation}
    \begin{aligned}
        \label{eq:ASPCA_Prime}
        \min_{ \textbf{w} \in \mathbb{R}^{d \times 1}} & \quad \| \textbf{A}- \textbf{y} \textbf{w}^{\top} \|_{2}^{2} + \lambda\|\textbf{w}\|_{1}\\
        s.t. & \quad \textbf{w}^{\top}\textbf{w} = 1,
    \end{aligned}
\end{equation}
where $\textbf{y} = \textbf{A}\textbf{w}$ is the approximation of the projected data $\textbf{A}\textbf{w}$. With such formulation, we are able to take advantage of temporarily treating $\textbf{y}$ as a known part when computing $\textbf{w}$, thus simplifying the optimization problem.

Similarly, we decompose $\| \textbf{A} - \textbf{y} \textbf{w}^{\top} \|_{2}^{2}$ into $K$ parts by
\begin{align}
    \| \textbf{A} - \textbf{y} \textbf{w}^{\top} \|_{2}^{2} = \sum_{i=1}^{K} \| \textbf{A}^{i} - \tilde{\textbf{y}}_{i} \textbf{w}^{\top} \|_{2}^{2},
\end{align}
where $\tilde{\textbf{y}}_{i} = \textbf{A}^{i}\textbf{w}$. Then, we can reformulate \eqref{eq:ASPCA_Prime} as a consensus optimization problem

\begin{equation}
    \begin{aligned}
        \label{eq:FASPCA_Prime}
        \min_{ \textbf{w}_{1},\cdots,\textbf{w}_{K} \in \mathbb{R}^{d \times 1}} & \quad \sum_{i=1}^{K} \| \textbf{A}^{i} - \textbf{y}_{i} \textbf{w}_{i}^{\top} \|_{2}^{2} + \lambda\|\textbf{z}\|_{1} \\
        s.t. & \quad \textbf{w}_{i}^{\top}\textbf{w}_{i} = 1 \text{ and } \textbf{w}_{i} = \textbf{z}, \forall  i,
    \end{aligned}
\end{equation}
where $\textbf{w}_{1},\cdots,\textbf{w}_{K}  \in \mathbb{R}^{d \times 1}$ are local parameter for each worker, $\textbf{z} \in \mathbb{R}^{d \times 1}$ is the consensus parameter, and $\textbf{y}_{i} = \textbf{A}^{i}\textbf{w}_{i},\ \forall i$, is the approximation of the projected data $\textbf{A}^{i}\textbf{w}_{i}$. In practice, the approximation is performed using the loading from the previous iteration.


By introducing dual variables $\textbf{u}_{i} \in \mathbb{R}^{d \times 1},\ i=1,\dots,k,$ for the equality constraint, we obtain the augmented Lagrangian,
\begin{equation}
    \begin{aligned}
        \label{eq:FASPCA_Lag}
        \mathcal{L}_{\rho}( \{ \textbf{w}_{i}, \textbf{u}_{i} \}_{i=1}^{K}, \textbf{z} ) = & \sum_{i=1}^{K} \left( \vphantom{\frac{\rho}{2}} \| \textbf{A}^{i} - \textbf{y}_{i} \textbf{w}_{i}^{\top} \|_{2}^{2} + \textbf{u}_{i}^{\top}(\textbf{w}_{i} - \textbf{z}) \right. \\
        &  \left. \vphantom{\| \textbf{A}^{i} - \textbf{y}_{i} \textbf{w}_{i}^{\top} \|_{2}^{2}} + \frac{\rho}{2} \| \textbf{w}_{i} - \textbf{z} \|_{2}^{2}\right) + \lambda\|\textbf{z}\|_{1},
    \end{aligned}
\end{equation}
where $\rho > 0 $ is a penalty parameter. The general ADMM algorithm for FASPCA is conducted in Algorithm \ref{alg:FASPCAP_Seudo} by minimizing $\mathcal{L}_{\rho}(\cdot)$ w.r.t. $ \textbf{w}_{i}, \textbf{z}, \textbf{u}_{i} $ alternately.
\begin{algorithm}
\caption{ADMM Algorithm for FASPCA}
\label{alg:FASPCAP_Seudo}
\begin{algorithmic}[1]
\Require
    Datasets $ \textbf{A}^{1},\cdots,\textbf{A}^{K}$, $\lambda$, $\rho$
\Ensure The sparse loading $\textbf{z}$
\State Initialize $\textbf{w}_{i}^{0}$ by a random normalized vector, $\forall i$,  $\textbf{u}_{i}^{0}, \textbf{z}^{0} = \textbf{0}, \forall i,$ and $t = 0$
\While{stopping criteria are not satisfied}

\noindent\texttt{Workers:} $// \ In\ parallel$ \Comment{local primal update} 
    \For{$i \in\{ 1,\dots,K\}$}
        \State $\textbf{y}_{i}^{t} \gets \textbf{A}^{i}\textbf{w}_{i}^{t}$
        \State $ \textbf{w}_{i}^{t+1} \gets \operatorname*{argmin}_{\textbf{w}_{i}^{\top}\textbf{w}_{i} = 1} \mathcal{L}_{\rho}( \{ \textbf{w}_{i}, \textbf{u}_{i}^{t} \}_{i=1}^{K}, \textbf{z}^{t} ) $
        \State Send $ \textbf{w}_{i}^{t+1} $ to the master
    \EndFor
    
\noindent\texttt{Master:}  \Comment{consensus primal update}
    \State $ \textbf{z}^{t+1} \gets \operatorname*{argmin}_{\textbf{z}} \mathcal{L}_{\rho}( \{ \textbf{w}_{i}^{t+1}, \textbf{u}_{i}^{t} \}_{i=1}^{K}, \textbf{z} ) $
    \State Send $ \textbf{z}^{t+1} $ to all workers
    
\noindent\texttt{Workers:} $// \ In\ parallel$ \Comment{local dual update}
    \For{$i \in\{ 1,\dots,K\}$}
        \State $ \textbf{u}_{i}^{t+1} \gets \textbf{u}_{i}^{t} + \rho(\textbf{w}_{i}^{t+1} - \textbf{z}^{t+1})$
    \EndFor
    \State $t \gets t + 1$
\EndWhile
\State $\textbf{z} \gets \textbf{z}^{t}$
\end{algorithmic}
\end{algorithm}

In the $\textbf{z}$-update phase, we replicate the approach used in problem \eqref{eq:FSSPCA_Prime}. This involves calculating the minimum of $ \mathcal{L}_{\rho}( \{ \textbf{w}_{i}, \textbf{u}_{i} \}_{i=1}^{K}, \textbf{z} ) $ with respect to $\textbf{z}$ by employing Algorithm \ref{alg:FSSPCA_z_update}, adhering to a consistent update pattern.

Next, let us introduce how to compute $\textbf{w}_{i}$ in Algorithm \ref{alg:FASPCAP_Seudo}. 
Given fixed $ \textbf{y}_{i} $, $ \textbf{u}_{i} $ and $\textbf{z}$, the optimization of $ \mathcal{L}_{\rho}( \{ \textbf{w}_{i}, \textbf{u}_{i} \}_{i=1}^{K}, \textbf{z} ) $ over $\textbf{w}_{i}$ is
\begin{equation}
    \begin{aligned}
        \label{alg:FASPCA_w_update}
        \min_{\textbf{w}_{i}\in\mathbb{R}^{d \times 1} } & \quad \| \textbf{A}^{i} - \textbf{y}_{i} \textbf{w}_{i}^{\top} \|_{2}^{2} +  \textbf{u}_{i}^{\top}\textbf{w}_{i} + \frac{\rho}{2} \| \textbf{w}_{i} - \textbf{z} \|_{2}^{2} \\
        s.t. & \quad \textbf{w}_{i}^{\top}\textbf{w}_{i} = 1.
    \end{aligned}
\end{equation}

Set $ f(\textbf{w}_{i}) := \| \textbf{A}^{i} - \textbf{y}_{i} \textbf{w}_{i}^{\top} \|_{2}^{2} +  \textbf{u}_{i}^{\top}\textbf{w}_{i} + \rho \| \textbf{w}_{i} - \textbf{z} \|_{2}^{2}/2$. To find the minimum of $f(\textbf{w}_{i})$, suppose $ \partial f(\textbf{w}_{i}) / \partial \textbf{w}_{i} = 0$, then we have 
\begin{align*}
    \textbf{w}_{i} = \frac{ 2(\textbf{A}^{i})^{\top} \textbf{y}_{i} - \textbf{u}_{i} + \rho\textbf{z} }{ 2\textbf{y}_{i}^{\top}\textbf{y}_{i} + \rho }.
\end{align*}
Note that since $(2\textbf{y}_{i}^{\top}\textbf{y}_{i} + \rho) \in \mathbb{R}^{1}$, we manage to keep the computational cost low in updating $\textbf{w}_{i}$. To fit the orthonormality constraint $\textbf{w}^{\top}\textbf{w} = 1$, we directly normalize $\textbf{w}_{i}$ as follows
\begin{align}
    \textbf{w}_{i} = \frac{\textbf{w}_{i}}{\| \textbf{w}_{i}\|_{2} }.
\end{align}

\begin{algorithm}
\caption{ Worker's Update for FASPCA}
\label{alg:ApproxFSPCA_wUpdate}
\begin{algorithmic}[1]
\Require
    $ \textbf{A}^{i}, \textbf{w}_{i}^{t}, \textbf{u}_{i}^{t}, \textbf{z}^{t}, \rho $
\Ensure $ \textbf{w}_{i}^{t+1} $
\State $\textbf{y}_{i}^{t} \gets \textbf{A}^{i} \textbf{w}^{t}$
\State${\textbf{w}}_{i}^{t+1} \gets \frac{ 2(\textbf{A}^{i})^{\top} \textbf{y}_{i}^{t} - \textbf{u}_{i}^{t} + \rho\textbf{z}^{t} }{2 (\textbf{y}_{i}^{t})^{\top} \textbf{y}_{i}^{t} + \rho }$
\State $ \textbf{w}_{i}^{t+1} \gets \frac{\textbf{w}_{i}^{t+1}}{\| \textbf{w}_{i}^{t+1} \|_{2} }$
\end{algorithmic}
\end{algorithm}

\subsection{Deflation method for FASPCA}
Similarly, we extend the FASPCA to compute the $r$th loading with the presence of the first $r-1$ loadings. The main distinction between FSSPCA and FASPCA is that the former computes multiple loadings simultaneously, while the latter computes one loading at a time. Therefore, we adapt Algorithm \ref{alg:SmoFSPCA_Defla} and modify Step 9 for FASPCA by:
\begin{align}
    \textbf{z}_{j} \gets \textbf{z}_{j}/\|\textbf{z}_{j}\|_{2}.
\end{align}

Suppose we have $r-1$ loadings $\{ \textbf{z}_{i} \}_{i=1}^{r-1}$ and intend to compute the $r$th loading, the problem is formulated as 
\begin{equation}
    \begin{aligned}
        \label{eq:FASPCA_Prime_Defla}
        \min_{ \textbf{w}_{1},\cdots,\textbf{w}_{K} \in \mathbb{R}^{d \times 1}} & \quad \sum_{i=1}^{K} \| (\textbf{A}^{i})^{D} - \textbf{y}_{i}^{D} \textbf{w}_{i}^{\top} \|_{2}^{2} + \lambda\|\textbf{z}\|_{1} \\
        s.t. & \quad \textbf{w}_{i}^{\top}\textbf{G}\textbf{w}_{i} = 1 \text{ and } \textbf{w}_{i} = \textbf{z}, \forall  i,
    \end{aligned}
\end{equation}
where $(\textbf{A}^{i})^{D} = \textbf{A}^{i}\textbf{G}$, $\textbf{y}_{i}^{D} = (\textbf{A}^{i})^{D}\textbf{w}_{i}$, and $\textbf{G} = \prod_{i=1}^{r-1} ( \textbf{I} - \textbf{z}_{i} \textbf{z}_{i}^{\top}) \in \mathbb{R}^{d \times d}$ lie in the orthogonal complement space of $\{ \textbf{z}_{i} \}_{i=1}^{r-1}$. The constraint $\textbf{w}_{i}^{\top}\textbf{G}\textbf{w}_{i} = 1$ ensures that the new loading $\textbf{z}_{r}$ must be orthogonal to the other loadings.

In this formulation, we can similarly find the minimum of $ \mathcal{L}_{\rho}^{D}( \{ \textbf{w}_{i}, \textbf{u}_{i} \}_{i=1}^{K}, \textbf{z} ) $ over $\textbf{z}$ using Algorithm \ref{alg:FSSPCA_z_update}. However, since the orthonormality constraint in \eqref{eq:FASPCA_Prime} and \eqref{eq:FASPCA_Prime_Defla} are slightly different, we modify the last step of workers' updates in Algorithm \ref{alg:ApproxFSPCA_wUpdate} as follows,
\begin{align}
    \label{eq:DeflOrth}
    \textbf{w}_{i} = \frac{\textbf{w}_{i}}{\| \textbf{G}\textbf{w}_{i}\|_{2} }.
\end{align}
Since $\textbf{G}$ is a orthogonal projection matrix, we have $\textbf{G} = \textbf{G}^{\top}$ and  $\textbf{G}^{\top}\textbf{G} = \textbf{G}$. Hence, by \eqref{eq:DeflOrth}, we can ensure $\textbf{w}_{i}^{\top}\textbf{G}\textbf{w}_{i} = 1$.

\section{Experiment}
In the federated learning framework, non-IID (non-Independently and Identically Distributed) data distribution across nodes is a critical concern, as it can significantly reduce both the accuracy and convergence rate \cite{kairouz2019advances, zhao2018federated}. Real-world scenarios often exhibit data skews and imbalances among participating workers due to a variety of factors such as user behavior, data collection methods, and other domain-specific influences. To evaluate the performance of our proposed methods, FSSPCA and FASPCA, we conducted experiments on four distinct datasets: IID synthetic data, non-IID synthetic data, IID real data, and non-IID real data. These evaluations were performed on a universally accessible laptop with an Intel Core i5-8250U CPU and 8 GB of RAM, operating without the assistance of a GPU. All algorithms were implemented using the Python programming environment.

\subsection{Experiments on IID synthetic data}
In the following experiments, we assess FSSPCA and FASPCA using synthetic data sampled from a zero-mean multivariate normal distribution with a covariance matrix $\boldsymbol{\Sigma}\in \mathbb{R}^{500 \times 500}$ that consists of sparse eigenvectors and split the dataset into $K$ parts to evaluate the performance of our methods in different distributed systems. Following \cite{journee2010generalized,tan2019learning}, we construct $\boldsymbol{\Sigma} = \textbf{V}\textbf{D}\textbf{V}^{\top}$, where $\textbf{V} \in \mathbb{R}^{500 \times 500}$ is an orthogonal matrix whose columns are eigenvectors of $\boldsymbol{\Sigma}$ and $\textbf{D} \in \mathbb{R}^{500 \times 500}$ is a diagonal matrix whose diagonal elements are eigenvalues of $\boldsymbol{\Sigma}$. We set the first two columns of \textbf{V} as the ground truth for sparse loadings, specifically setting the first 10 elements of the first column and the second 10 elements of the second column to $\frac{1}{\sqrt{10}}$, while all other elements are set to 0. The remaining columns of $\textbf{V}$ are random samples from a uniform distribution over [0, 1). The first two eigenvalues of $\textbf{D}$ are set to 400 and 300, respectively, and the rest are set to 1. Our aim is to recover the first two significant eigenvectors, denoted as $\textbf{V}{sol}$, and performance is evaluated by the \textit{recovery error} $\epsilon = \| \textbf{z}\textbf{z}^{\top} - \textbf{V}_{sol}\textbf{V}_{sol}^{\top} \|_{F}^{2}$, where $\textbf{z}$ is the outcome of our methods.

\begin{figure}[htbp]
    \centering
    \subfloat[FSSPCA]{
        \includegraphics[width=0.3\linewidth]{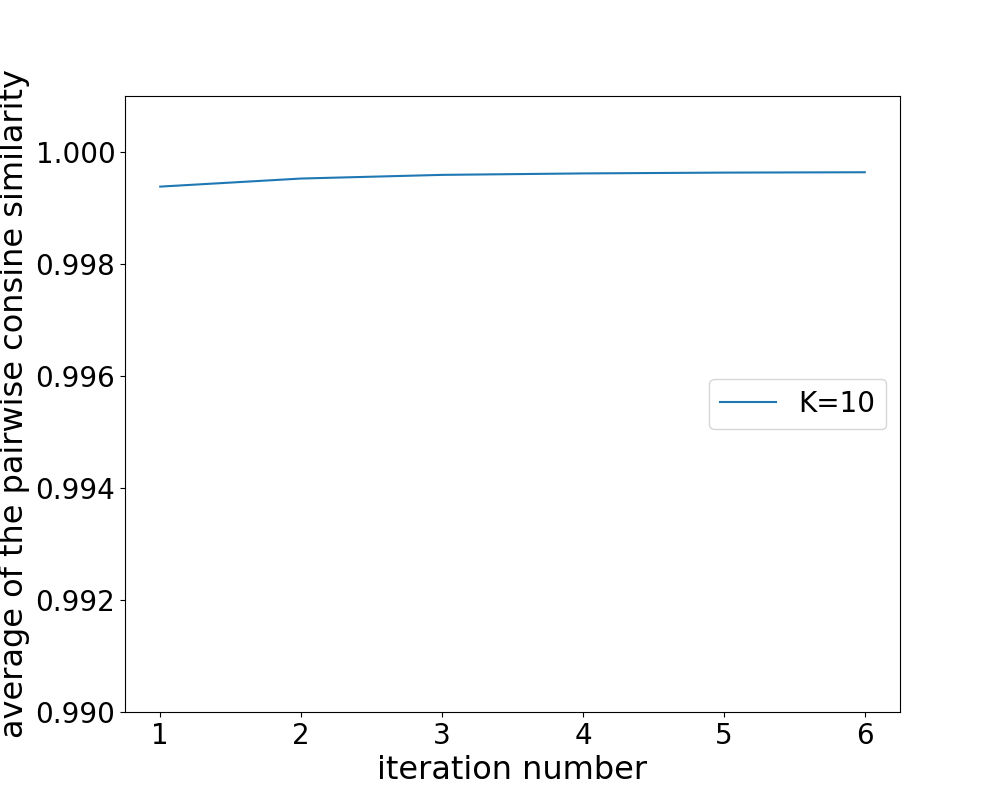}
        \label{fig:FSSPCA_Synthetic_worker_similarity_second_K_10_L1_500_L2_300}
    }
    \hfill 
    \subfloat[FASPCA \nth{1} loading]{
        \includegraphics[width=0.3\linewidth]{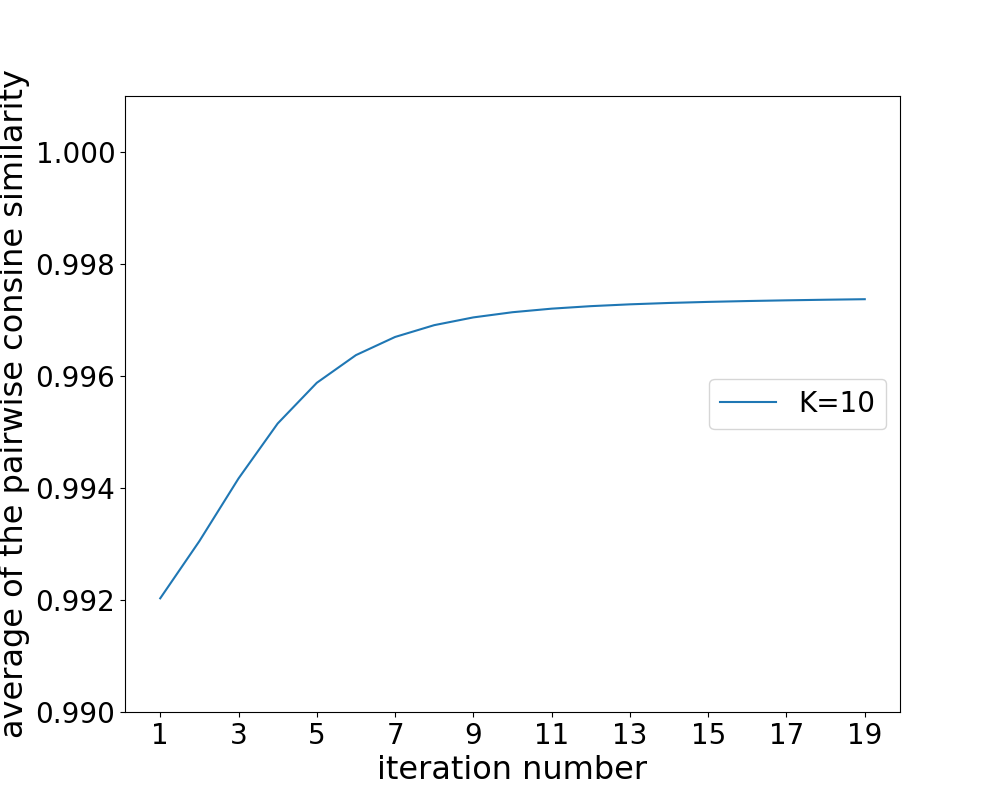}
        \label{fig:FASPCA_Synthetic_worker_similarity_first_K_10_L_1000}
    }
    \hfill 
    \subfloat[FASPCA \nth{2} loading]{
        \includegraphics[width=0.3\linewidth]{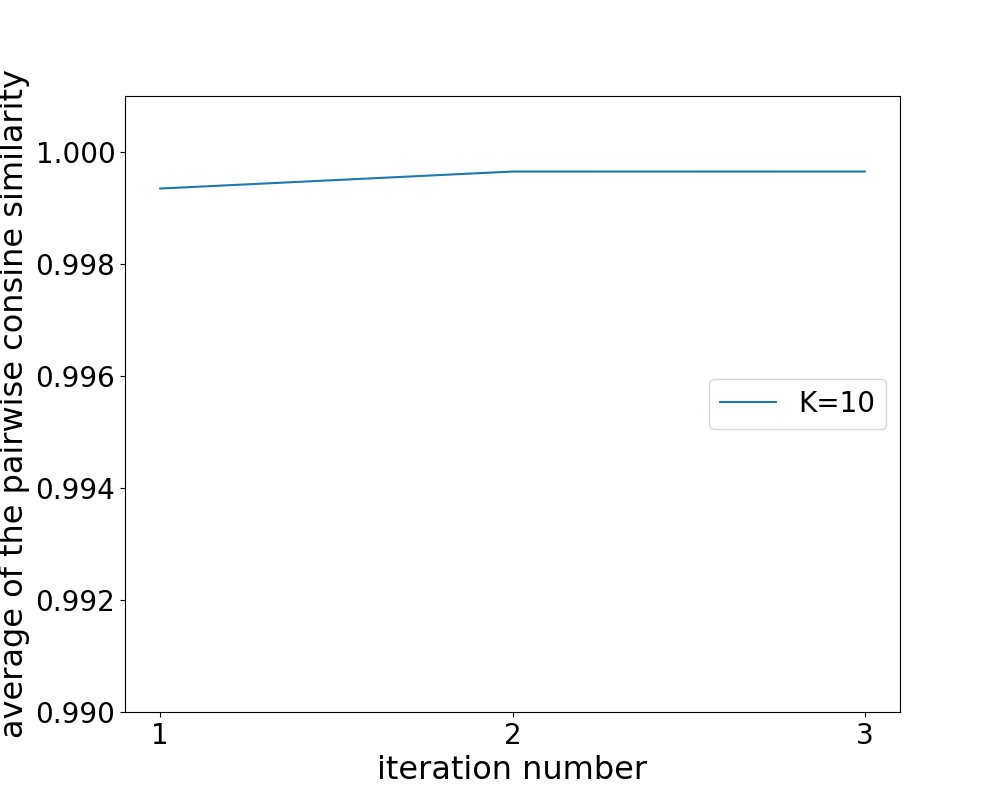}
        \label{fig:FASPCA_Synthetic_worker_similarity_second_K_10_L_1000}
    }
    
    \caption{The similarity performance of all workers for the first two loadings on the synthetic data.The settings of FSSPCA are $\lambda_{1} = 50$, $\lambda_{2} = 100$ and $\rho=1000$. The settings of FASPCA are $\lambda = 50$ and $\rho=1000$. }
    \label{fig:FASPCA_Synthetic_similarity}
\end{figure}

Firstly, we demonstrate that parameters from each worker become increasingly similar with each iteration to attain a consensus model parameter. To evaluate this, we set $K=10$ and compute the pairwise cosine similarity of all workers' parameters, and then average their absolute values. Fig. \ref{fig:FASPCA_Synthetic_similarity} illustrates that with an increasing number of iterations, the average value approaches  1. In other words, the parameters from each worker will become more and more similar to each other, which satisfies the constraint of the prime problem for FASPCA \eqref{eq:FASPCA_Prime} and FSSPCA \eqref{eq:FSSPCA_Prime}.

\begin{figure}[htbp]
    \centering
    \subfloat[FSSPCA]{
        \includegraphics[width=0.3\linewidth]{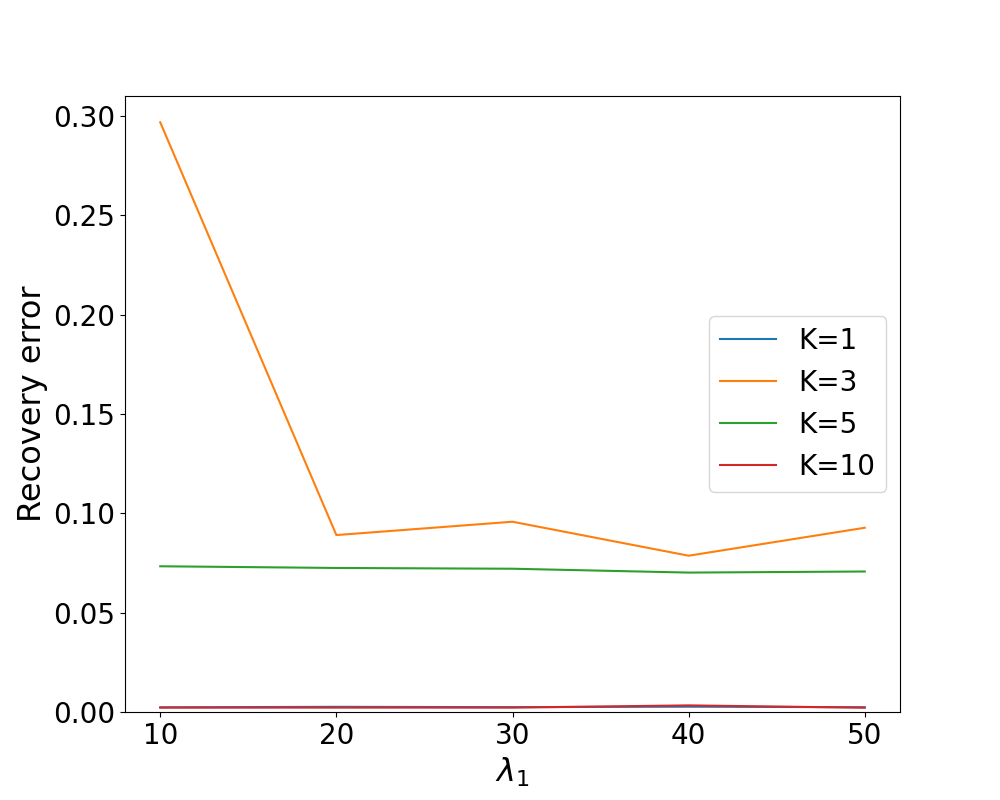}
    \label{fig:FSSPCA_Synthetic_recovery_error_second_K}
    }
    \hfill 
    \subfloat[FASPCA \nth{1} loading.]{
        \includegraphics[width=0.3\linewidth]{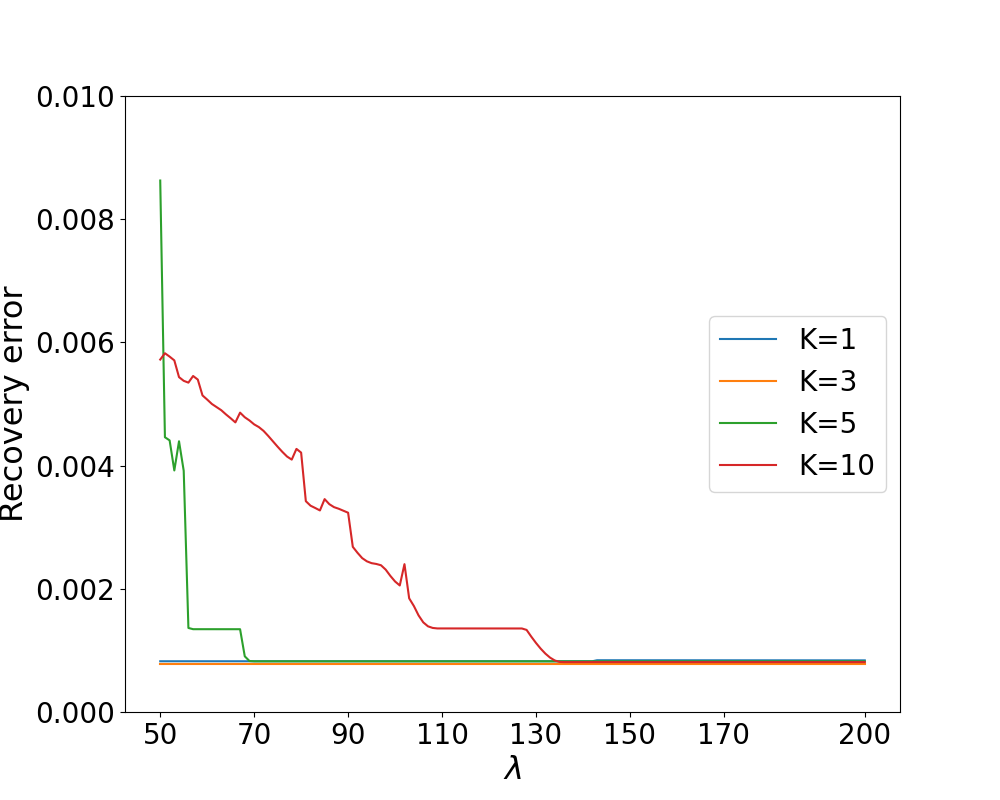}
    \label{fig:FASPCA_Synthetic_recovery_error_first_K}
    }
    \hfill 
    \subfloat[FASPCA \nth{2} loading]{
        \includegraphics[width=0.3\linewidth]{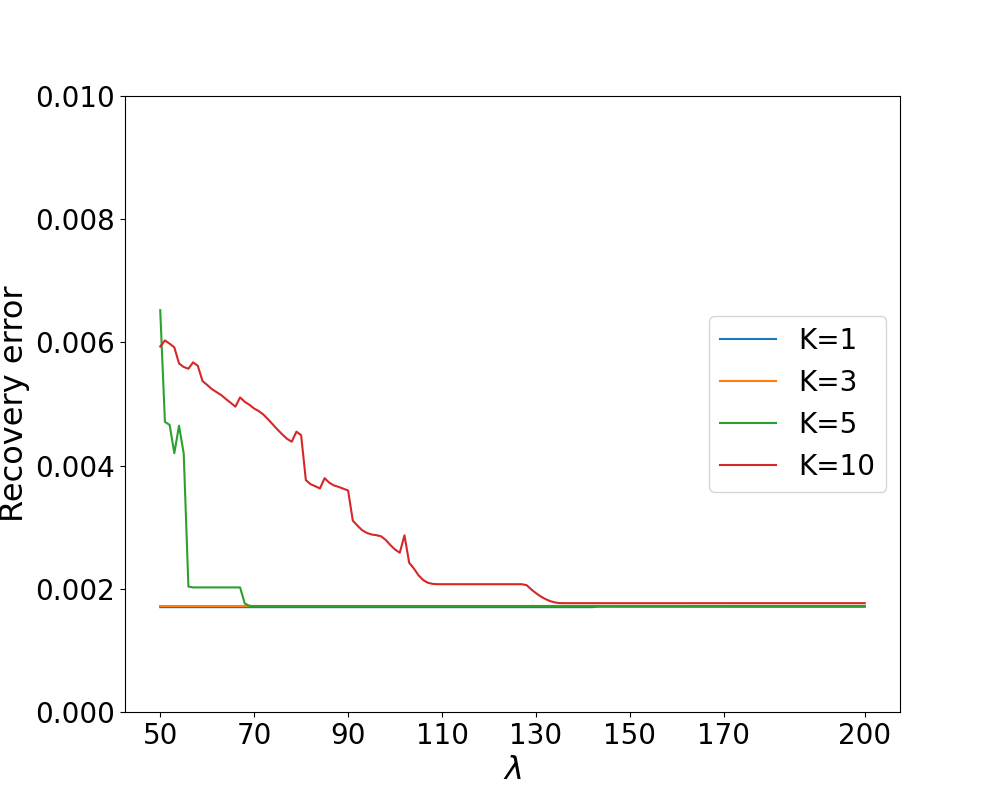}
    \label{fig:FASPCA_Synthetic_recovery_error_second_K}
    }

    \caption{The recovery error performance with different $\lambda$ or $\lambda_{1}$ of workers for the first two loadings on the synthetic data when $K=1,3,5,10$. The settings of FSSPCA are $\lambda_{1}$ from 10 to 50, $\lambda_{2} = 300$ and $\rho=1000$. The settings of FASPCA are $\lambda$ from 50 to 200 and $\rho=1000$. }
    \label{fig:FASPCA_Synthetic_recovery_error_K}
\end{figure}

Next, we split the synthetic dataset into $K$ parts to evaluate the recovery error of our methods with different numbers of data owners, where $K = 1,3,5,10$. As shown in Fig. \ref{fig:FASPCA_Synthetic_recovery_error_K}, regardless of the value of $K$, as the $\lambda$ or $\lambda_{1}$ increase, the recovery error approaches 0. In other words, our methods yield outcomes that closely resemble the ground truth $\textbf{V}_{sol}$. Hence, we conclude that our proposed models FASPCA and FSSPCA are capable of obtaining sparse loadings under the FL framework.

\subsection{Experiments on IID Real data}
Wisconsin Diagnostic Breast Cancer Dataset (WDBC)  \cite{dua2019} is a well-known dataset for the classification task. This dataset contains 569 instances with 31 features. The primary goal of SPCA is to identify important features. To evaluate the performance of our proposed models, we choose the last 30 features of WDBC, excluding the ID number. We augmented it with 800 randomly generated features from a uniform distribution over [0, 1) to create a new dataset $\text{WDBC}^{\star}\in\mathbb{R}^{569 \times 830}$. In the following experiments, our goal is to identify 2 sparse loadings. We demonstrate that our proposed models can effectively identify important features by shrinking the weights of the added random features to zero or close to zero.

\begin{figure}[htbp]
  \centering
  \subfloat[$\lambda_{1} = 0$.]{
    \includegraphics[width=0.45\linewidth]{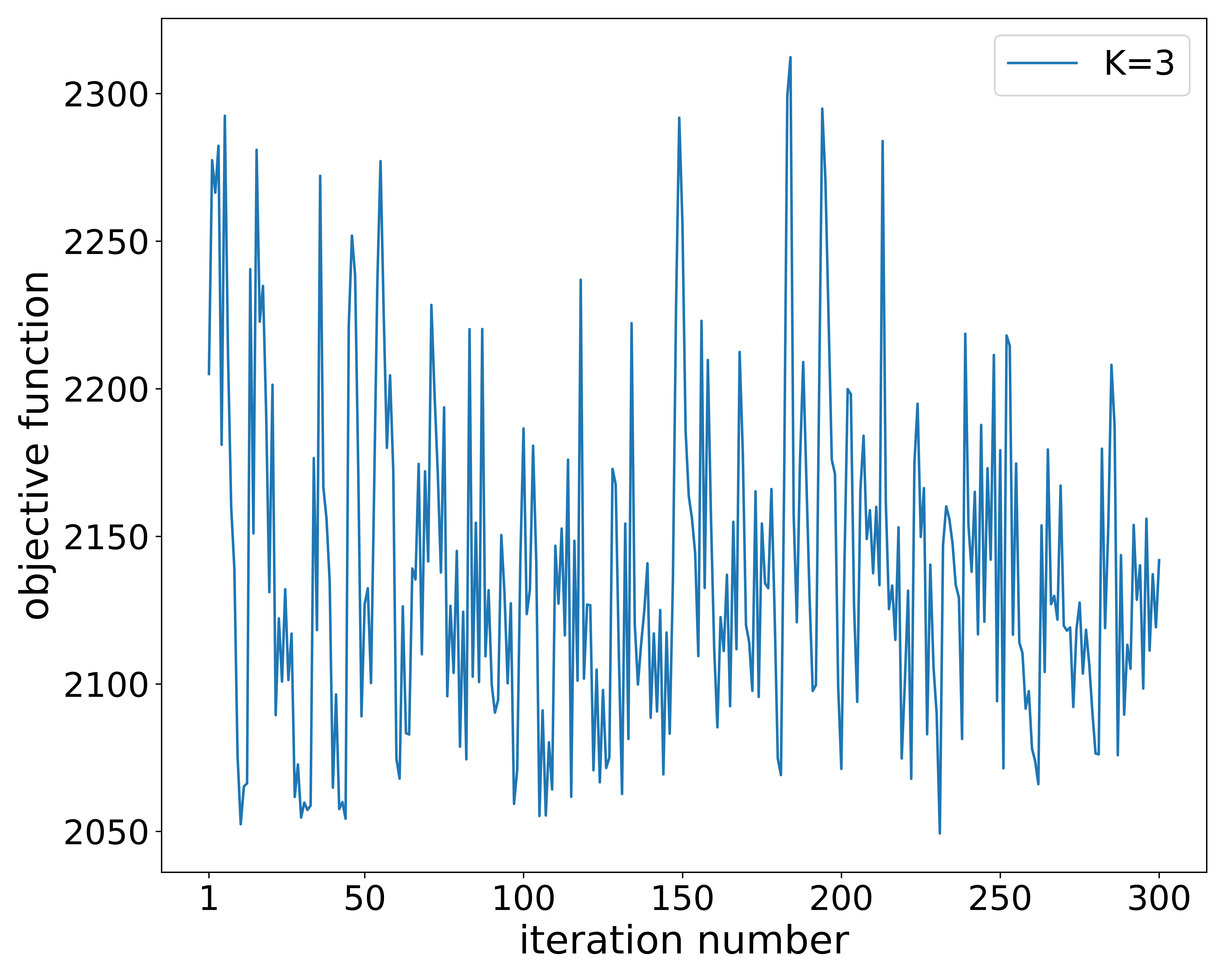}
    \label{fig:FSSPCA_WDBC_counter_L_0}
  }
  \subfloat[$\lambda_{1} = 10$.]{
    \includegraphics[width=0.45\linewidth]{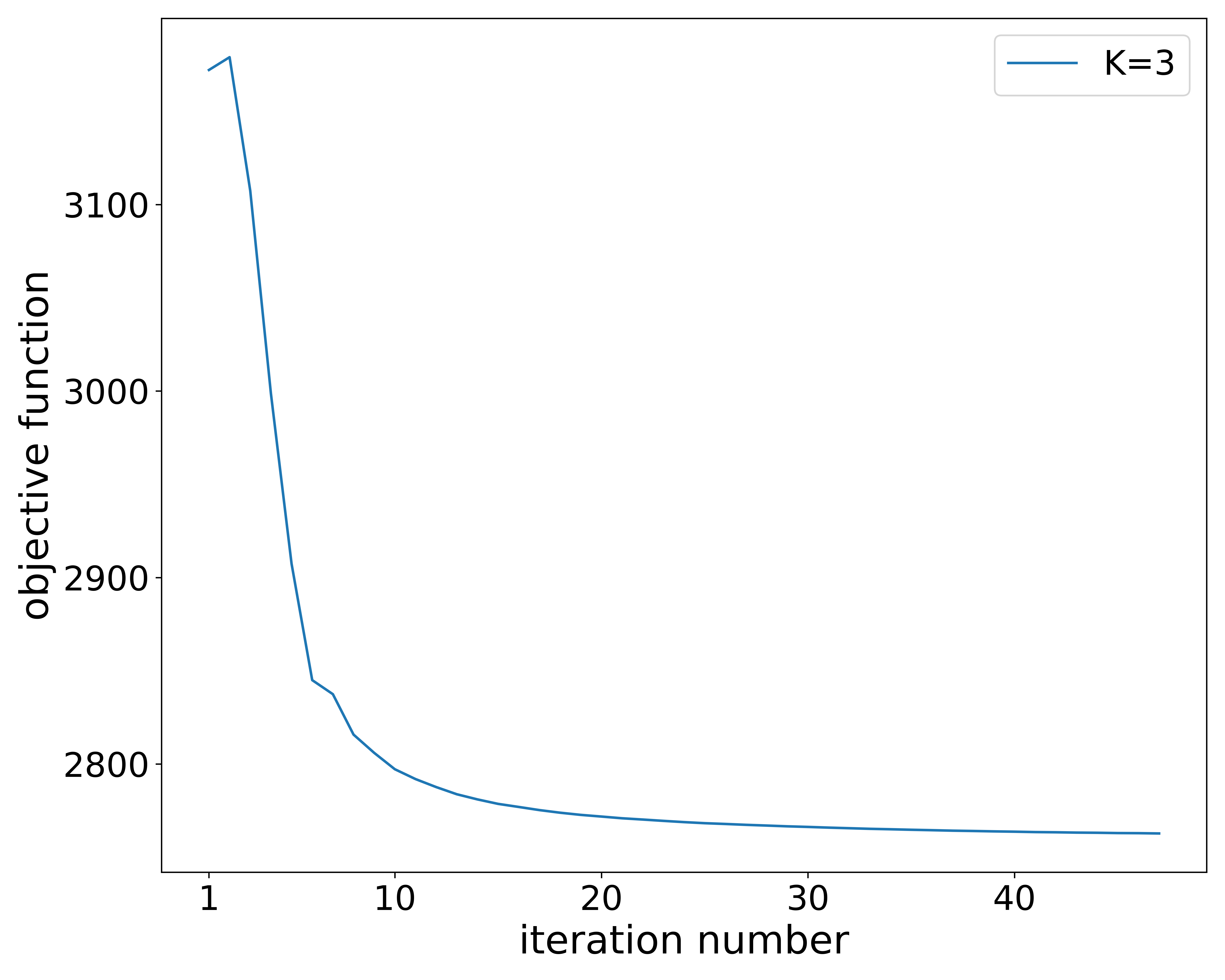}
    \label{fig:FSSPCA_WDBC_counter_L_10}
  }
  \caption{FSSPCA: the objective function changes on $\text{WDBC}^{\star}$ data, with $\lambda_{1}\in\{0,10\}$ and $\lambda_{2}=30$.}
  \label{fig:FSSPCA_WDBC_counter_L}
\end{figure}

Firstly, we show the effectiveness of adding the $\ell_{1}$-norm smoothing terms for FSSPCA. As shown in Fig. \ref{fig:FSSPCA_WDBC_counter_L}, when $\lambda_{1}=10$, which means that we activate the smoothing function term $r(\textbf{w}_{i})$, the convergence behavior is far better than the case of $\lambda_{1}=0$. That is to say, adding the $\ell_{1}$-norm smoothing terms significantly improves convergence. 

\begin{figure}[htbp]
  \centering
    \subfloat[FSSPCA]{
    \includegraphics[width=0.45\linewidth]{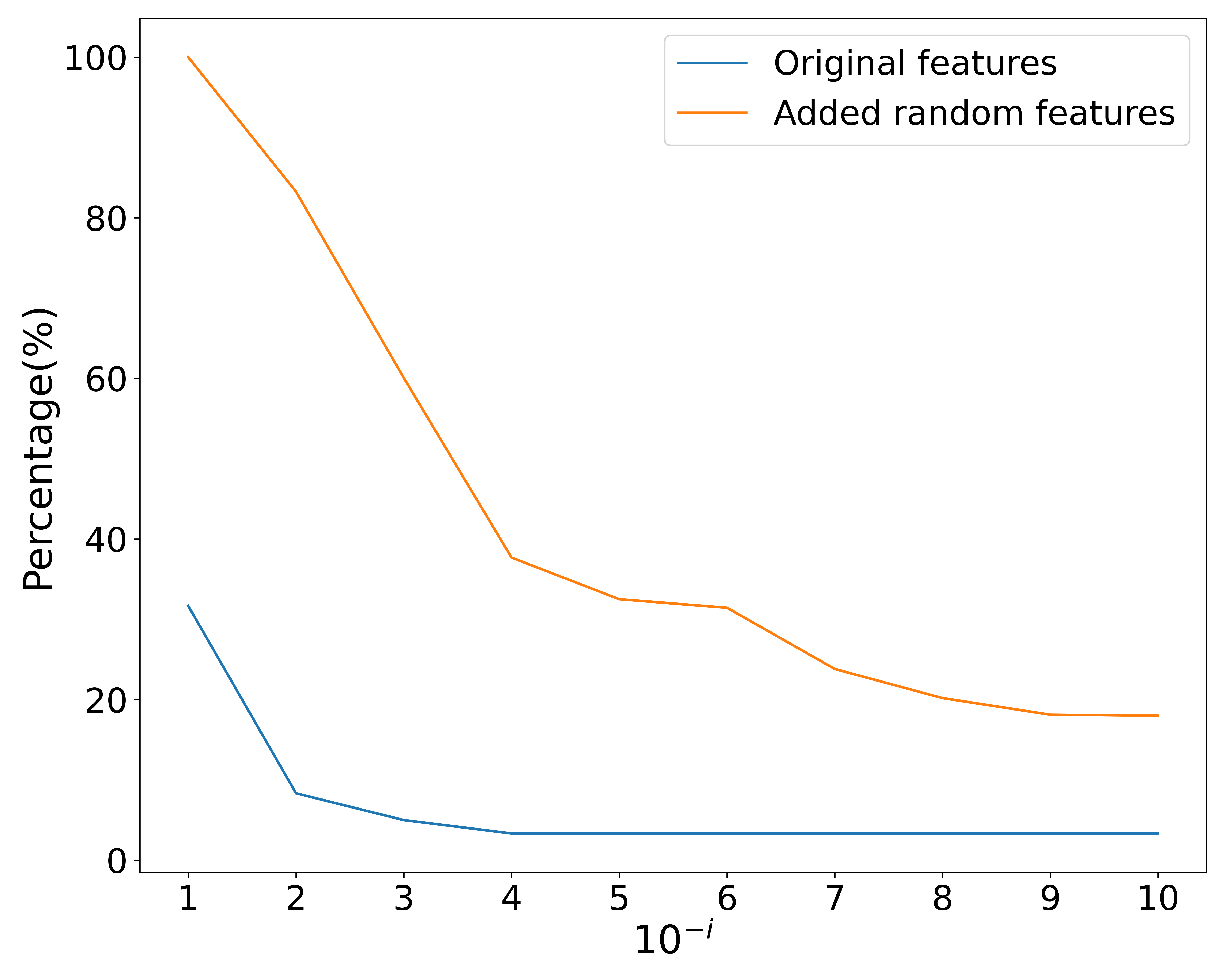}
    \label{fig:FSSPCA_WDBC_small_L_10}
  }
  \subfloat[FASPCA]{
    \includegraphics[width=0.45\linewidth]{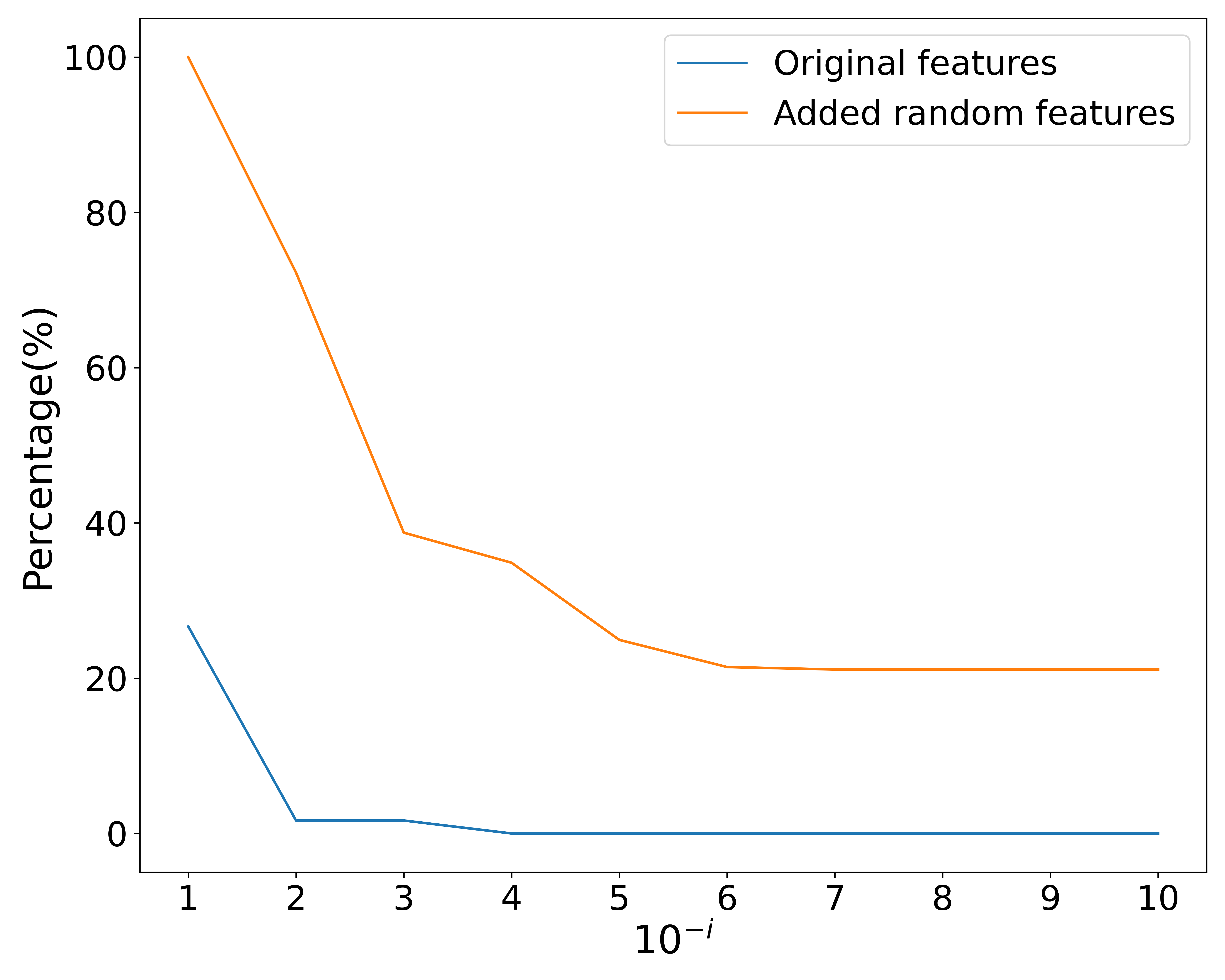}
    \label{fig:FASPCA_WDBC_small_L_60}
  }

  \caption{The comparison of the percentage of small values, falling within $[-10^{-i},10^{-i}]$, between original features and added random features.The settings of FSSPCA are $\lambda_{1}=10$, $\lambda_{2} = 30$ and $\rho=1000$. The settings of FASPCA are $\lambda=60$ and $\rho=1000$. }
  \label{fig:FSSPC_WDBC_small_percentage}
\end{figure}

Since one of the primary purposes of SPCA is to facilitate interpretability through sparse loadings, we assess the result of the purposed methods. In Fig. \ref{fig:FSSPC_WDBC_small_percentage}, for both FASPCA and FSSPCA, we observed that all the weights associated with the added random features lie in $[-0.1,0.1]$. Additionally, 27$\%$ and 35$\%$ of the weights corresponding to original features of FASPCA and FSSPCA fell within $[-0.1,0.1]$, respectively. Moreover, for $i\in\{ 1,2,\cdots,10 \}$, we noticed that the percentages of small values, falling within $[-10^{i},10^{i}]$, of added random features are higher than the original features. These findings confirm that both FASPCA and FSSPCA provide more interpretable results and can effectively identify important features in the dataset.

As shown in Table \ref{tab:WDBC_performance}, no matter FASPCA or FSSPCA, the reconstruction errors of our proposed models are close to PCA and the $\ell_{0}$-norm of our proposed models are less than PCA. Thus, our proposed model can indeed attain loading sparsity. Additionally, while FASPCA is faster in terms of computation time, FSSPCA requires fewer iterations, almost half that of FASPCA. In our experiments, we assume negligible transmission costs between workers and the master. However, in scenarios with significant transmission costs, FSSPCA may offer advantages over FASPCA.

\begin{table}[htbp]
    \caption{Performance comparison with PCA, FASPCA, and FSSPCA on $\text{WDBC}^{\star}$ data. The settings of FASPCA are $\lambda=170$ and $\rho=1000$. The settings of FSSPCA are $\lambda_{1}=10$, $\lambda_{2} = 190$ and $\rho=1000$.}
    \begin{center}
        \begin{tabular}{|c|c|c|c|c|}
        \hline
        \textbf{Method} & \multicolumn{1}{c|}{\textbf{Reconstruction Error}} & \textbf{$\ell_{0}$-norm} & \textbf{Time (sec)} & \textbf{Iteration} \\
        \hline
        PCA & 677.3092 & 1660 & - & - \\
        \hline
        FASPCA & 680.2175 & 483 & \textbf{7.8995} & 315 \\
        \hline
        FSSPCA & 680.9359 & 629 & 36.4278 & \textbf{133} \\
        \hline
        \end{tabular}
    \end{center}
    \label{tab:WDBC_performance}
\end{table}
\subsection{Experiments on non-IID synthetic data}
In the third experiment, we evaluate our proposed model on non-IID synthetic data. The data is generated in a manner similar to the IID synthetic dataset, with eigenvalues and the first two eigenvectors remaining consistent across all workers. However, for each worker, the remaining eigenvectors are uniquely generated as random samples from a normal distribution $N(0, \sigma_i)$, and $\sigma_i$ is independently sampled from a uniform distribution over [0, 1) for each worker.

\begin{figure}[htbp]
    \centering
    \subfloat[FSSPCA]{
        \includegraphics[width=0.3\linewidth]{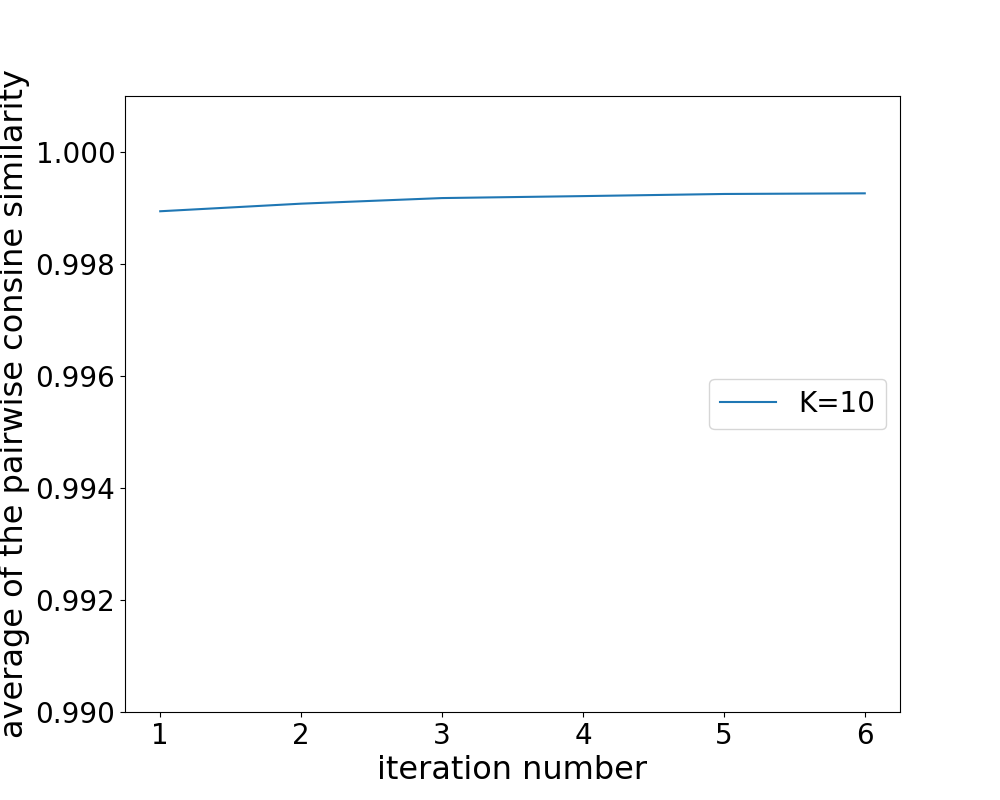}
    }
    \hfill 
    \subfloat[FASPCA \nth{1} loading]{
        \includegraphics[width=0.3\linewidth]{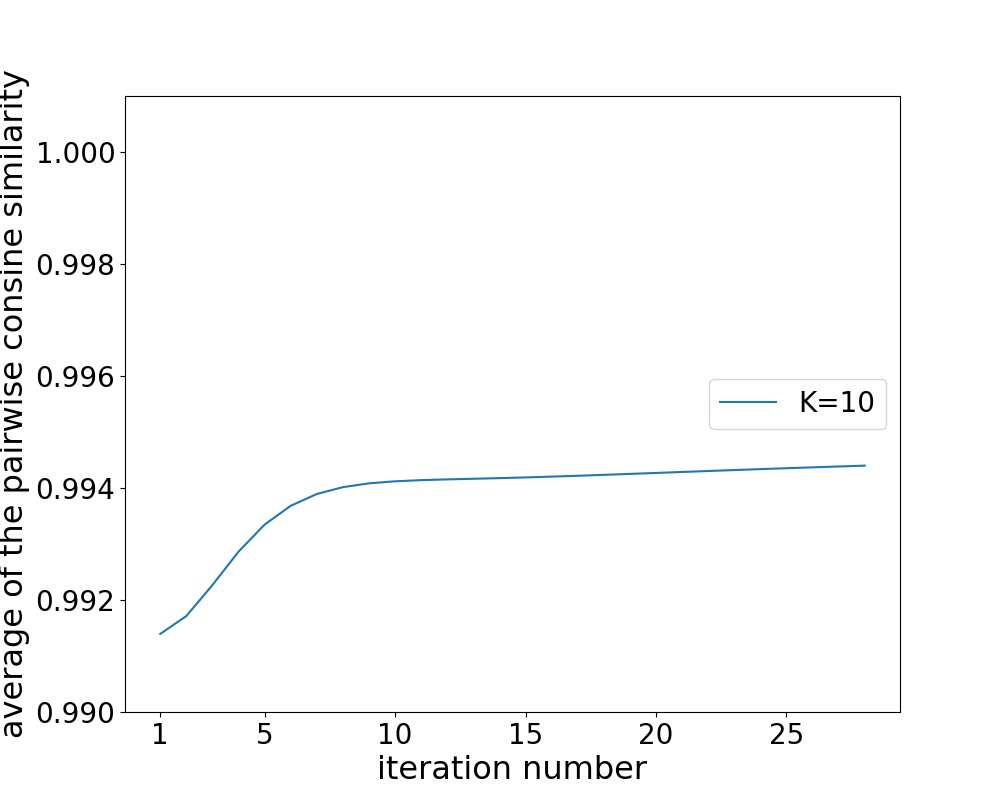}
    }
    \hfill 
    \subfloat[FASPCA \nth{2} loading]{
        \includegraphics[width=0.3\linewidth]{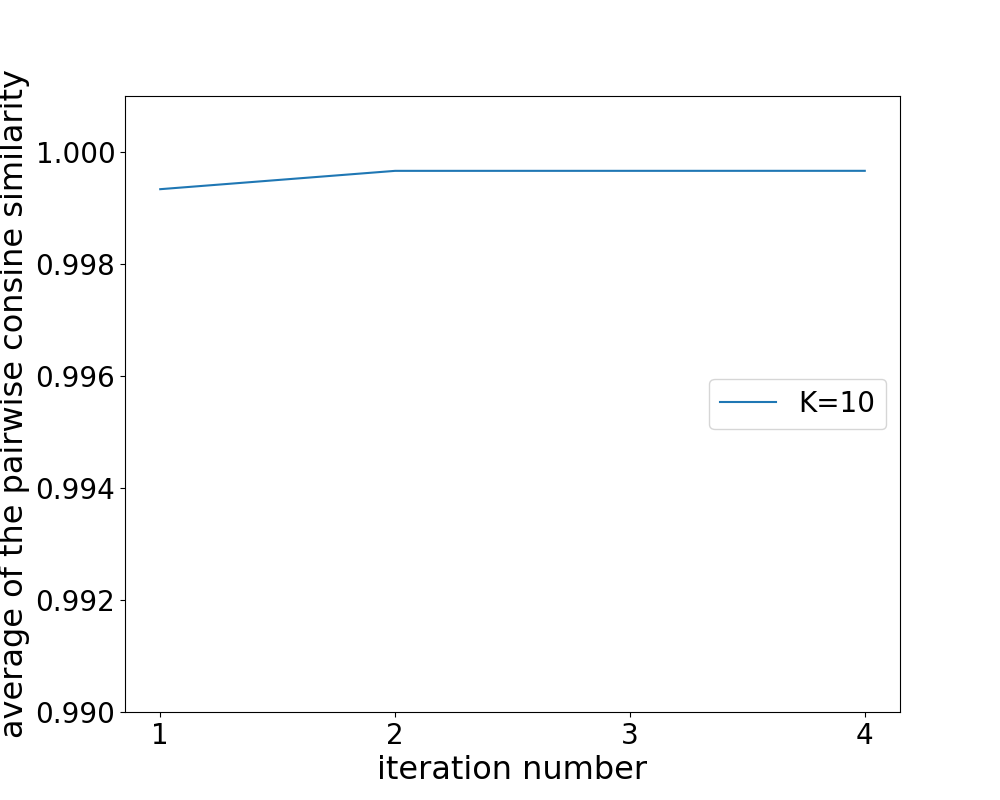}
    }

    \caption{The similarity performance of all workers for the first two loadings on the synthetic data. The settings of FSSPCA are $\lambda_{1} = 50$, $\lambda_{2} = 100$ and $\rho=1000$.The settings of FASPCA are $\lambda = 50$ and $\rho=1000$. }
    \label{fig:Synthetic_noniid_similarity}
\end{figure}

As demonstrated in Fig. \ref{fig:Synthetic_noniid_similarity}, the trends observed in non-IID synthetic data are consistent with those in IID synthetic data. Specifically, as the number of iterations increases, the average value converges toward 1. This pattern demonstrates that the parameters from each worker become increasingly aligned, thereby meeting the constraints of the primary problems as defined for both FASPCA \eqref{eq:FASPCA_Prime} and FSSPCA \eqref{eq:FSSPCA_Prime}. Moreover, the recovery errors for FASPCA and FSSPCA were 0.0855 and 0.02466, respectively. Both of which are close to zero, indicating an effective approach towards the ground truth $\textbf{V}_{sol}$. The robustness of our proposed methods in both IID and non-IID settings highlights their effectiveness and broad applicability within the federated learning framework.

\subsection{Experiments on non-IID Real data}
In the fourth experiment, we assess the performance of our proposed model on non-IID real data. Similar to Subsection B, we retain the last 30 features of the WDBC dataset, but introduce modifications to the additional 800 random features. These 800 features are now uniquely allocated to each participating worker. For each worker, the allocated features are generated using a normal distribution with a mean of zero and a variance sampled from a uniform distribution over [0, 1). The remaining features are rendered sparse, with 80\% being exactly 0 and 20\% non-zero values generated from a uniform distribution over [0, 1). With this generating scheme, each worker may return the local model with some non-zero weights associated with the allocated parts, while the other workers returning sparse results. 

\begin{figure}[htbp]
  \centering
  \subfloat[FSSPCA]{
    \includegraphics[width=0.45\linewidth]{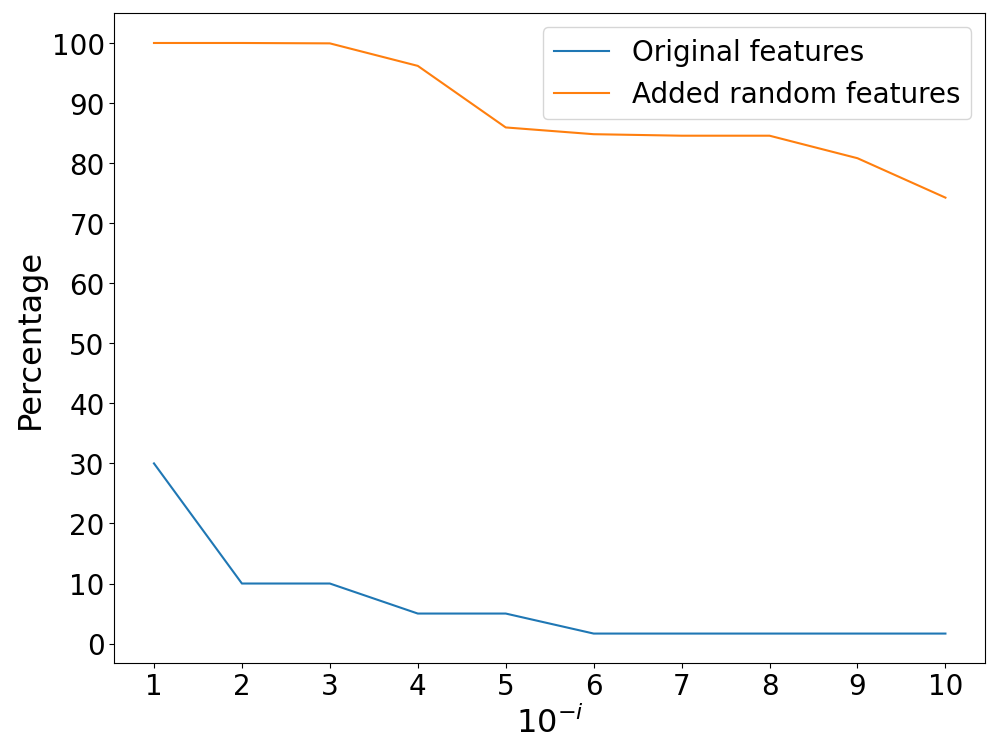}
  }
  \subfloat[FASPCA]{
    \includegraphics[width=0.45\linewidth]{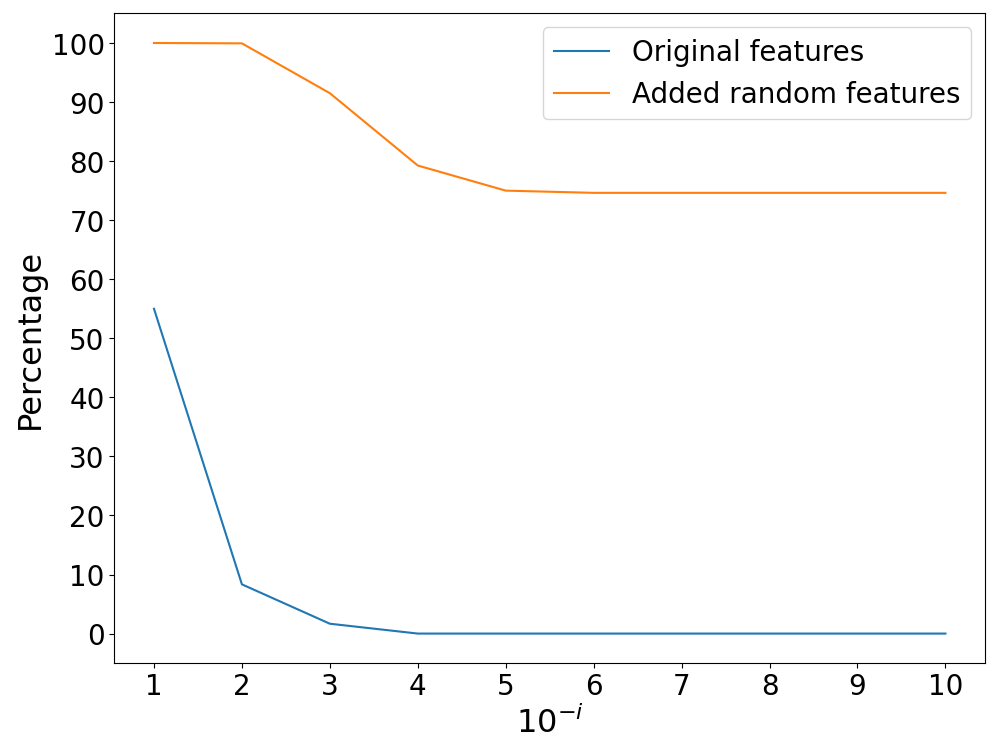}
  }
  
  \caption{The comparison of the percentage of small values, which lies in $[-10^{-i},10^{-i}]$ between original features and added random features. The settings of FSSPCA are $\lambda_{1}=10$, $\lambda_{2} = 30$ and $\rho=1000$.The settings of FASPCA are $\lambda=60$ for first loading, $\lambda=600$ for second loading and $\rho=1000$. }
  \label{fig:WDBC_noniid_small}
\end{figure}

Referring to Fig. \ref{fig:WDBC_noniid_small}, it is evident that for both FASPCA and FSSPCA, the weights of the added random features fall within the range of $[-0.01, 0.01]$. In contrast, 55$\%$ of the original feature weights for FASPCA and 30$\%$ for FSSPCA lie within the $[-0.1, 0.1]$ range. Moreover, for $i$ in the set $\{1,2,\ldots,10\}$, the proportions of smaller values that fall within the bounds of $[-10^{i},10^{i}]$ are notably higher for the added random features compared with the original features. Another critical aspect to consider is the reconstruction error. The reconstruction error for PCA itself stands at 236.4406, while for FASPCA and FSSPCA, it stands at 242.4147 and 236.9857, respectively. These close values indicate that both FASPCA and FSSPCA perform comparably to traditional PCA. These results show that both FASPCA and FSSPCA work well in identifying important features and provide easily interpretable outcomes, even when applied to the non-IID dataset.

\section{Conclusion and Future Work}

We have introduced two innovative approaches: Federated Approximation Sparse Principal Component Analysis (FASPCA) and Federated Smoothing Sparse Principal Component Analysis (FSSPCA). These methods are designed to tackle the Sparse Principal Component Analysis (SPCA) challenge within a distributed framework. They enable model training on client-side devices, preserving privacy by keeping data localized. Instead of transmitting raw data to a central server, only model updates are exchanged, significantly enhancing data security. We have also incorporated the least squares approximation in the PCA component to reduce computational complexity without compromising accuracy. In specific scenarios, such as anomaly detection, where only a single loading is required, FASPCA demonstrates remarkable efficiency due to its rapid computation. When multiple loadings are necessary, the deflation technique is employed to map the data to the orthogonal complement space of the leading component. Our experiments, conducted on synthetic and $\text{WDBC}^{\star}$ datasets, underscore the exceptional performance of both FASPCA and FSSPCA in obtaining sparse loadings within distributed systems. Furthermore, these methods effectively extract crucial features while accommodating non-IID (non-Independently and Identically Distributed) random features.

In our future research endeavors, we intend to explore the integration of cryptographic techniques, such as differential privacy, to further enhance data security and privacy within the federated learning framework.



\printbibliography[title={References}]

@inproceedings{lee2019CRSVM,
  author={Chen, Hsiang-Hsuan and Lee, Yuh-Jye},
  booktitle={2019 IEEE International Conference on Big Data (Big Data)}, 
  title={Distributed Consensus Reduced Support Vector Machine}, 
  year={2019},
  volume={},
  number={},
  pages={5718-5727},
  doi={10.1109/BigData47090.2019.9006098}}

@inproceedings{mcmahan2017communication,
  title={Communication-efficient learning of deep networks from decentralized data},
  author={McMahan, Brendan and Moore, Eider and Ramage, Daniel and Hampson, Seth and y Arcas, Blaise Aguera},
  booktitle={Artificial Intelligence and Statistics},
  pages={1273--1282},
  year={2017},
  organization={PMLR}
}

@article{kairouz2019advances,
  title={Advances and open problems in federated learning},
  author={Kairouz, Peter and McMahan, H Brendan and Avent, Brendan and Bellet, Aur{\'e}lien and Bennis, Mehdi and Bhagoji, Arjun Nitin and Bonawitz, Keith and Charles, Zachary and Cormode, Graham and Cummings, Rachel and others},
  journal={arXiv preprint arXiv:1912.04977},
  year={2019}
}

@article{zhao2018federated,
  title={Federated learning with non-iid data},
  author={Zhao, Yue and Li, Meng and Lai, Liangzhen and Suda, Naveen and Civin, Damon and Chandra, Vikas},
  journal={arXiv preprint arXiv:1806.00582},
  year={2018}
}

@article{pearson1901liii,
  title={{LIII.} On lines and planes of closest fit to systems of points in space},
  author={Pearson, Karl},
  journal={The London, Edinburgh, and Dublin Philosophical Magazine and Journal of Science},
  volume={2},
  number={11},
  pages={559--572},
  year={1901},
  publisher={Taylor \& Francis}
}

@article{grammenos2020federated,
  title={Federated Principal Component Analysis},
  author={Grammenos, Andreas and Mendoza Smith, Rodrigo and Crowcroft, Jon and Mascolo, Cecilia},
  journal={Advances in Neural Information Processing Systems},
  volume={33},
  year={2020}
}

@article{zou2018selective,
  title={A selective overview of sparse principal component analysis},
  author={Zou, Hui and Xue, Lingzhou},
  journal={Proceedings of the IEEE},
  volume={106},
  number={8},
  pages={1311--1320},
  year={2018},
  publisher={IEEE}
}

@article{jolliffe2003modified,
  title={A modified principal component technique based on the {LASSO}},
  author={Jolliffe, Ian T and Trendafilov, Nickolay T and Uddin, Mudassir},
  journal={Journal of computational and Graphical Statistics},
  volume={12},
  number={3},
  pages={531--547},
  year={2003},
  publisher={Taylor \& Francis}
}

@article{zou2006sparse,
  title={Sparse principal component analysis},
  author={Zou, Hui and Hastie, Trevor and Tibshirani, Robert},
  journal={Journal of computational and graphical statistics},
  volume={15},
  number={2},
  pages={265--286},
  year={2006},
  publisher={Taylor \& Francis}
}

@article{journee2010generalized,
  title={Generalized power method for sparse principal component analysis.},
  author={Journ{\'e}e, Michel and Nesterov, Yurii and Richt{\'a}rik, Peter and Sepulchre, Rodolphe},
  journal={Journal of Machine Learning Research},
  volume={11},
  number={2},
  year={2010}
}

@inproceedings{ge2018minimax,
  title={Minimax-optimal privacy-preserving sparse pca in distributed systems},
  author={Ge, Jason and Wang, Zhaoran and Wang, Mengdi and Liu, Han},
  booktitle={International Conference on Artificial Intelligence and Statistics},
  pages={1589--1598},
  year={2018},
  organization={PMLR}
}

@inproceedings{vu2013fantope,
  title={Fantope projection and selection: A near-optimal convex relaxation of sparse {PCA}},
  author={Vu, Vincent Q and Cho, Juhee and Lei, Jing and Rohe, Karl},
  booktitle={Advances in neural information processing systems},
  pages={2670--2678},
  year={2013}
}

@article{tan2019learning,
  title={Learning sparse {PCA} with stabilized {ADMM} method on stiefel manifold},
  author={Tan, Mingkui and Hu, Zhibin and Yan, Yuguang and Cao, Jiezhang and Gong, Dong and Wu, Qingyao},
  journal={IEEE Transactions on Knowledge and Data Engineering},
  year={2019},
  publisher={IEEE}
}

@article{ma2013alternating,
  title={Alternating direction method of multipliers for sparse principal component analysis},
  author={Ma, Shiqian},
  journal={Journal of the Operations Research Society of China},
  volume={1},
  number={2},
  pages={253--274},
  year={2013},
  publisher={Springer}
}

@inproceedings{hajinezhad2015nonconvex,
  title={Nonconvex alternating direction method of multipliers for distributed sparse principal component analysis},
  author={Hajinezhad, Davood and Hong, Mingyi},
  booktitle={2015 IEEE Global Conference on Signal and Information Processing (GlobalSIP)},
  pages={255--259},
  year={2015},
  organization={IEEE}
}

@article{lee2012anomaly,
  title={Anomaly detection via online oversampling principal component analysis},
  author={Lee, Yuh-Jye and Yeh, Yi-Ren and Wang, Yu-Chiang Frank},
  journal={IEEE transactions on knowledge and data engineering},
  volume={25},
  number={7},
  pages={1460--1470},
  year={2012},
  publisher={IEEE}
}

@article{papadimitriou2005streaming,
  title={Streaming pattern discovery in multiple time-series},
  author={Papadimitriou, Spiros and Sun, Jimeng and Faloutsos, Christos},
  year={2005},
  publisher={Carnegie Mellon University}
}

@article{yang1995projection,
  title={Projection approximation subspace tracking},
  author={Yang, Bin},
  journal={IEEE Transactions on Signal processing},
  volume={43},
  number={1},
  pages={95--107},
  year={1995},
  publisher={IEEE}
}

@inproceedings{mackey2008deflation,
  title={Deflation Methods for Sparse {PCA}.},
  author={Mackey, Lester W},
  booktitle={NIPS},
  volume={21},
  pages={1017--1024},
  year={2008}
}

@article{glowinski1975approximation,
  title={Sur l'approximation, par {\'e}l{\'e}ments finis d'ordre un, et la r{\'e}solution, par p{\'e}nalisation-dualit{\'e} d'une classe de probl{\`e}mes de Dirichlet non lin{\'e}aires},
  author={Glowinski, Roland and Marroco, A},
  journal={ESAIM: Mathematical Modelling and Numerical Analysis-Mod{\'e}lisation Math{\'e}matique et Analyse Num{\'e}rique},
  volume={9},
  number={R2},
  pages={41--76},
  year={1975}
}

@article{gabay1976dual,
  title={A dual algorithm for the solution of nonlinear variational problems via finite element approximation},
  author={Gabay, Daniel and Mercier, Bertrand},
  journal={Computers \& mathematics with applications},
  volume={2},
  number={1},
  pages={17--40},
  year={1976},
  publisher={Elsevier}
}

@book{boyd2011distributed,
  title={Distributed optimization and statistical learning via the alternating direction method of multipliers},
  author={Boyd, Stephen and Parikh, Neal and Chu, Eric},
  year={2011},
  publisher={Now Publishers Inc}
}

@inproceedings{damgaard2012multiparty,
  title={Multiparty computation from somewhat homomorphic encryption},
  author={Damg{\aa}rd, Ivan and Pastro, Valerio and Smart, Nigel and Zakarias, Sarah},
  booktitle={Annual Cryptology Conference},
  pages={643--662},
  year={2012},
  organization={Springer}
}

@inproceedings{cramer2001multiparty,
  title={Multiparty computation from threshold homomorphic encryption},
  author={Cramer, Ronald and Damg{\aa}rd, Ivan and Nielsen, Jesper B},
  booktitle={International conference on the theory and applications of cryptographic techniques},
  pages={280--300},
  year={2001},
  organization={Springer}
}

@inproceedings{garay2003strengthening,
  title={Strengthening zero-knowledge protocols using signatures},
  author={Garay, Juan A and MacKenzie, Philip and Yang, Ke},
  booktitle={International Conference on the Theory and Applications of Cryptographic Techniques},
  pages={177--194},
  year={2003},
  organization={Springer}
}

@article{kingma2014adam,
  title={Adam: A method for stochastic optimization},
  author={Kingma, Diederik P and Ba, Jimmy},
  journal={arXiv preprint arXiv:1412.6980},
  year={2014}
}

@article{he2018cola,
  title={Cola: Decentralized linear learning},
  author={He, Lie and Bian, An and Jaggi, Martin},
  journal={arXiv preprint arXiv:1808.04883},
  year={2018}
}

@article{smith2018cocoa,
  title={{CoCoA}: A general framework for communication-efficient distributed optimization},
  author={Smith, Virginia and Forte, Simone and Chenxin, Ma and Tak{\'a}{\v{c}}, Martin and Jordan, Michael I and Jaggi, Martin},
  journal={Journal of Machine Learning Research},
  volume={18},
  pages={230},
  year={2018},
  publisher={MIT press}
}

@inproceedings{silva2019federated,
  title={Federated learning in distributed medical databases: Meta-analysis of large-scale subcortical brain data},
  author={Silva, Santiago and Gutman, Boris A and Romero, Eduardo and Thompson, Paul M and Altmann, Andre and Lorenzi, Marco},
  booktitle={2019 IEEE 16th international symposium on biomedical imaging (ISBI 2019)},
  pages={270--274},
  year={2019},
  organization={IEEE}
}

@inproceedings{zheng2019helen,
  title={Helen: Maliciously secure coopetitive learning for linear models},
  author={Zheng, Wenting and Popa, Raluca Ada and Gonzalez, Joseph E and Stoica, Ion},
  booktitle={2019 IEEE Symposium on Security and Privacy (SP)},
  pages={724--738},
  year={2019},
  organization={IEEE}
}

@article{lee2001ssvm,
  title={{SSVM}: A smooth support vector machine for classification},
  author={Lee, Yuh-Jye and Mangasarian, Olvi L},
  journal={Computational optimization and Applications},
  volume={20},
  number={1},
  pages={5--22},
  year={2001},
  publisher={Springer}
}

@article{lee2005spl,
  title={$\epsilon$-{SSVR}: a smooth support vector machine for $\epsilon$-insensitive regression},
  author={Lee, Yuh-Jye and Hsieh, Wen-Feng and Huang, Chien-Ming},
  journal={IEEE Transactions on knowledge and data engineering},
  volume={17},
  number={5},
  pages={678--685},
  year={2005},
  publisher={IEEE}
}

@article{hebiri2011smooth,
  title="The Smooth-Lasso and other $\ell$1+ $\ell$2-penalized methods",
  author={Hebiri, Mohamed and Van De Geer, Sara and others},
  journal={Electronic Journal of Statistics},
  volume={5},
  pages={1184--1226},
  year={2011},
  publisher={The Institute of Mathematical Statistics and the Bernoulli Society}
}

@article{saheya2018numerical,
  title={Numerical comparisons based on four smoothing functions for absolute value equation},
  author={Saheya, B and Yu, Cheng-He and Chen, Jein-Shan},
  journal={Journal of Applied Mathematics and Computing},
  volume={56},
  number={1},
  pages={131--149},
  year={2018},
  publisher={Springer}
}

@article{saheya2019neural,
  title={Neural network based on systematically generated smoothing functions for absolute value equation},
  author={Saheya, B and Nguyen, Chieu Thanh and Chen, Jein-Shan},
  journal={Journal of Applied Mathematics and Computing},
  volume={61},
  number={1},
  pages={533--558},
  year={2019},
  publisher={Springer}
}

@book{absil2009optimization,
  title={Optimization algorithms on matrix manifolds},
  author={Absil, P-A and Mahony, Robert and Sepulchre, Rodolphe},
  year={2009},
  publisher={Princeton University Press}
}

@article{ring2012optimization,
  title={Optimization methods on Riemannian manifolds and their application to shape space},
  author={Ring, Wolfgang and Wirth, Benedikt},
  journal={SIAM Journal on Optimization},
  volume={22},
  number={2},
  pages={596--627},
  year={2012},
  publisher={SIAM}
}

@article{sato2019cholesky,
  title={Cholesky {QR}-based retraction on the generalized Stiefel manifold},
  author={Sato, Hiroyuki and Aihara, Kensuke},
  journal={Computational Optimization and Applications},
  volume={72},
  number={2},
  pages={293--308},
  year={2019},
  publisher={Springer}
}

@misc{dua2019,
author = "Dua, Dheeru and Graff, Casey",
year = "2017",
title = "{UCI} Machine Learning Repository",
url = "http://archive.ics.uci.edu/ml",
institution = "University of California, Irvine, School of Information and Computer Sciences" }


\end{document}